\documentclass{article}

\usepackage[english]{babel}

\usepackage{amsmath}
\usepackage{graphicx}
\usepackage[colorlinks=true, allcolors=blue]{hyperref}
\usepackage{arxiv}
\usepackage{tabularx}
\usepackage[utf8]{inputenc} 
\usepackage[T1]{fontenc}    
\usepackage{hyperref}       
\usepackage{url}            
\usepackage{booktabs}       
\usepackage{amsfonts}       
\usepackage{nicefrac}       
\usepackage{microtype}      
\usepackage{lipsum}
\graphicspath{ {./images/} }
\usepackage{booktabs}
\usepackage{multirow}
\usepackage{siunitx}

\begin{document}
\title{PGCS: Physical Law embedded Generative Cloud Synthesis in remote sensing images}

\author{
 Liying Xu \\
  School of Resource and Environmental Sciences\\
  Wuhan University\\
  Wuhan 430079, China \\
  \texttt{liyingxuwhu@whu.edu.cn} \\
   \And
 Huifang Li \\
  School of Resource and Environmental Sciences\\
  Wuhan University\\
  Wuhan 430079, China \\
  \texttt{huifangli@whu.edu.cn} \\
  \And
 Huanfeng Shen \\
  School of Resource and Environmental Sciences\\
  Wuhan University\\
  Wuhan 430079, China \\
  \texttt{shenhf@whu.edu.cn} \\
  \And
 Mingyang Lei \\
  School of Resource and Environmental Sciences\\
  Wuhan University\\
  Wuhan 430079, China \\
  \texttt{leimingyang@whu.edu.cn} \\
  \And
 Tao Jiang \\
  School of Resource and Environmental Sciences\\
  Wuhan University\\
  Wuhan 430079, China \\
  \texttt{jiangta0@whu.edu.cn} \\
}

\maketitle

\begin{abstract}
Data quantity and quality are both critical for information extraction and analyzation in remote sensing. However, the current remote sensing datasets often fail to meet these two requirements, for which cloud is a primary factor degrading the data quantity and quality. This limitation affects the precision of results in remote sensing application, particularly those derived from data-driven techniques. In this paper, a physical law embedded generative cloud synthesis method (PGCS) is proposed to generate diverse realistic cloud images to enhance real data and promote the development of algorithms for subsequent tasks, such as cloud correction, cloud detection, and data augmentation for classification, recognition, and segmentation. The PGCS method involves two key phases: spatial synthesis and spectral synthesis. In the spatial synthesis phase, a style-based generative adversarial network is utilized to simulate the spatial characteristics, generating an infinite number of single-channel clouds. In the spectral synthesis phase, the atmospheric scattering law is embedded through a local statistics and global fitting method, converting the single-channel clouds into multi-spectral clouds. The experimental results demonstrate that PGCS achieves a high accuracy in both phases and performs better than three other existing cloud synthesis methods. Two cloud correction methods are developed from PGCS and exhibits a superior performance compared to state-of-the-art methods in the cloud correction task. Furthermore, the application of PGCS with data from various sensors was investigated and successfully extended. Code will be provided at \url{https://github.com/Liying-Xu/PGCS}.
\end{abstract}

\section{Introduction}
\setlength{\parindent}{2em}
\noindent\hspace{2em} Remote sensing technology is an essential source of information for understanding and analyzing processes on the Earth's surface. The increasing availability of remote sensing platforms has led to a vast amount of observational data, which has significantly advanced the development of applications in this field. However, the quality and quantity of data affect the performance of algorithms on remote sensing tasks.

Clouds are frequently regarded as a major factor contributing to data degradation. The significant radiative differences between cloudy and cloud-free regions in remote sensing images pose challenges for accurately extracting land surface information for subsequent analysis. Therefore, various cloud removal methods have been proposed to improve the radiative quality of remote sensing images. The available cloud removal methods can be broadly divided into two main categories, i.e., model driven methods \cite{RN18,RN33,RN10} and data driven methods. Within the data driven methods, there are traditional statistical methods \cite{RN11,RN12,RN37,RN36} and deep learning based methods \cite{RN44,RN19,RN67,RN73}. Owing to their strong advantage in learning nonlinear relationships, the deep learning based methods have been demonstrated to be superior in cloud removal, particularly when dealing with complex cloud degradation. These algorithms are rapidly advancing, but they might face limitations from real-world data, potentially leading to a performance plateau \cite{RN89}. 

Take cloud removal as an example, the periodic observations by remote sensing sensors enable the construction of paired “cloudy \& cloud-free” datasets from real world. Researchers often assume that two remote sensing images acquired in close temporal proximity contain the same land surface information. Hence, images taken under different atmospheric conditions but within adjacent time phases can be selected to build such datasets. Some researchers \cite{RN73} have built datasets based on Landsat-8 data, which capture images in a period of 16 days. Meraner, et al. \cite{RN67} acquired paired cloudy and cloud-free Sentinel-2 images within the same meteorological season to construct the dataset, thereby ensuring that the paired data exhibited limited surface changes. Xu, et al. \cite{RN75} created a dataset using EO-1 Hyperion images, comprising 43 pairs of hyperspectral images captured at the same location and on close dates.

However, this multi-temporal approach exhibits three significant limitations. Firstly, the time intervals between consecutive images often exceed 10 days. Such lengthy intervals make it improbable that the surface conditions will have remained unchanged, particularly in dynamic environments such as agricultural regions during the growth season, flood-prone areas, and urban zones with dense human activity. As a result, the constructed datasets may fail to support the recovery of real land surface information. Secondly, the wide variation in cloud formations makes it difficult to automatically collect paired data for all cloud types, especially thin clouds with a diffuse distribution and indistinct boundaries. The manual selection process leads to the collection of paired data being both time-intensive and laborious. Thirdly, the limited number of available datasets often results in sample imbalance issues, adversely affecting the performance of cloud removal methods.

Therefore, developing data synthesis methods is crucial for sustaining progress in the field of remote sensing. Cloud synthesis methods can generate diverse and realistic cloud images, which enabling the creation of cloudy data with accurate label to address these limitations. When the label is clouds related data, it is instrumental in advancing techniques such as cloud removal, cloud correction and cloud detection. By correcting cloud interference and recovering the true land surface information, these methods significantly enhance data quality. Moreover, cloud synthesis as a data augmentation technique can enhance the capability of algorithms such as identification, classification, recognition, and segmentation to extract accurate information from low-quality data. This approach increases the quantity of available data for remote sensing tasks. 

The existing cloud synthetic methods can be broadly categorized into two main types: computer generation based methods and cloud retrieval based methods. Methods based on computer generation employ Perlin noise to simulate clouds without relying on real reference data. In 1985, Ken Perlin first introduced the concept of “solid texture” to create highly convincing representations of clouds \cite{RN58}. Building on this concept, Enomoto, et al. \cite{RN7} synthesized cloudy images by combining them with clear images using alpha blending, successfully applying this technique to the task of cloud removal. Subsequently, Czerkawski, et al. \cite{RN61} developed a novel approach to control the generation of synthetic clouds. However, these methods are entirely visual simulations, while ignoring the spectral characteristics of cloud in remote sensing images, which results in low-quality synthetic data.

Methods based on cloud retrieval estimate the cloud features extracted from actual cloudy images and then combine real cloud-free images to generate new synthetic cloudy images, which are paired with the real cloud-free images. For instance, Pan, et al. \cite{RN62} employed the dark channel prior to estimate the transmission map of real cloudy images. Subsequently, they implemented an inverse dehazing process to integrate this transmission map with a cloud-free image, effectively synthesizing a paired cloudy image. Zi, et al. \cite{RN37} extracted a reference cloud thickness map and a set of thickness coefficients from real cloudy images and then proposed a wavelength-dependent cloud simulation method based on an additive model. In contrast, Xu, et al. \cite{RN63} proposed a cloud self-subtraction operation to extract cloud components from real cloudy images captured over sea regions. However, due to their heavy reliance on real cloudy data, these methods cannot synthesize an infinite number of cloudy images, limiting the quantity of synthetic data. 

The quality and quantity of the synthetic data are also essential. However, the existing methods are unable to achieve these two goals simultaneously. To address this challenge, physical law embedded generative cloud synthesis (PGCS) is proposed in this paper, PGCS is a hybrid modeling approach that effectively leverages the respective strengths of the data-driven learning and physical process models. Generative artificial intelligence enables the production of an infinite variety of high-fidelity outputs in the spatial domain, and the physical law provides accurate descriptions of the radiative transfer process, which is essential for precise spectral simulation. The code for the cloud synthesis method and dataset construction is provided at \url{https://github.com/Liying-Xu/PGCS}.

\section{Related Work}
\label{sec:2}
\subsection{Scattering law for clouds}
\label{sec:2.1}
\setlength{\parindent}{2em}
\noindent\hspace{2em} Chavez Jr \cite{RN4} conducted a detailed analysis to quantitatively model the intensity of the hybrid scattering occurring in clouds. The cloud radiation $C$ is related to the wavelength $\lambda$ and atmospheric conditions $D$, and can be expressed as follows:
\begin{equation}
C = \frac{D}{\lambda ^ \gamma}\label{eq:eq1}
\end{equation}
\noindent where $\gamma$ is a parameter ranging from 0 to 4, depending on the level of atmospheric turbidity.

Chavez Jr \cite{RN4}  conducted an analysis and statistical study, which yielded five specific values for parameter $\gamma$, i.e., 4, 2, 1, 0.7, and 0.5. These values correspond to five distinct atmospheric conditions, i.e., very clear, clear, moderate, hazy, and very hazy, respectively. 

At the moment of imaging in remote sensing, the atmospheric conditions within the same pixel are generally consistent. Therefore, parameters $D$ and $\gamma$ can be assumed to be the same across the different spectral bands for a pixel. Consequently, the cloud radiation of the different spectral bands was related, and the relationship was constructed as shown in Equation (\ref{eq:eq2}), based on images captured by Landsat-8/9 \cite{RN31}. Parameter $\gamma$ can be estimated by utilizing the linear relationship between the coastal and blue bands in cloud-free regions, with the support of the cirrus band.
\begin{equation}
C_j = (\frac{\lambda_i}{\lambda_j})^\gamma \cdot C_i\label{eq:eq2}
\end{equation}

\noindent where $i$ and $j$ are a pair of band indices between the coastal, visible and near-infrared (VNIR) and cirrus bands.

However, this method is limited by its reliance on the cirrus band and clear pixels, which restricts its applicability. Furthermore, the pixel-by-pixel computation is complex, leading to low computational efficiency and pixel-wise spatial fragmentation. Therefore, it is necessary to propose a more applicable computational method to solve the key parameter of the scattering law and make the integration of the data and physical law robust.

\subsection{StyleGAN}
\label{sec:2.2}
\setlength{\parindent}{2em}
\noindent\hspace{2em} The Style-Based Generator Architecture (StyleGAN), which is a revolutionary image synthesis approach, was first released in 2019 by NVIDIA Corporation researchers \cite{RN64,RN79}. StyleGAN has since found applications in various fields, including image generation \cite{RN99,RN98}, data augmentation \cite{RN101,RN100}, and semantic image editing \cite{RN96,RN97}. StyleGAN excels in capturing both global structure and local details, resulting in visually appealing and coherent outputs. With its ability to generate diverse and unique variations, StyleGAN expands the creative possibilities beyond mere replication. However, there exists a research gap regarding the application of StyleGAN to remote sensing images, particularly in quantitative applications. To address this gap, it is crucial to develop methodologies that enable the generation of synthetic images with physical mechanisms, thereby enhancing their usability in quantitative analyses and evaluations.

\section{Methodology}
\label{sec:3}
\begin{figure}
\centering
\includegraphics[width=0.7\linewidth]{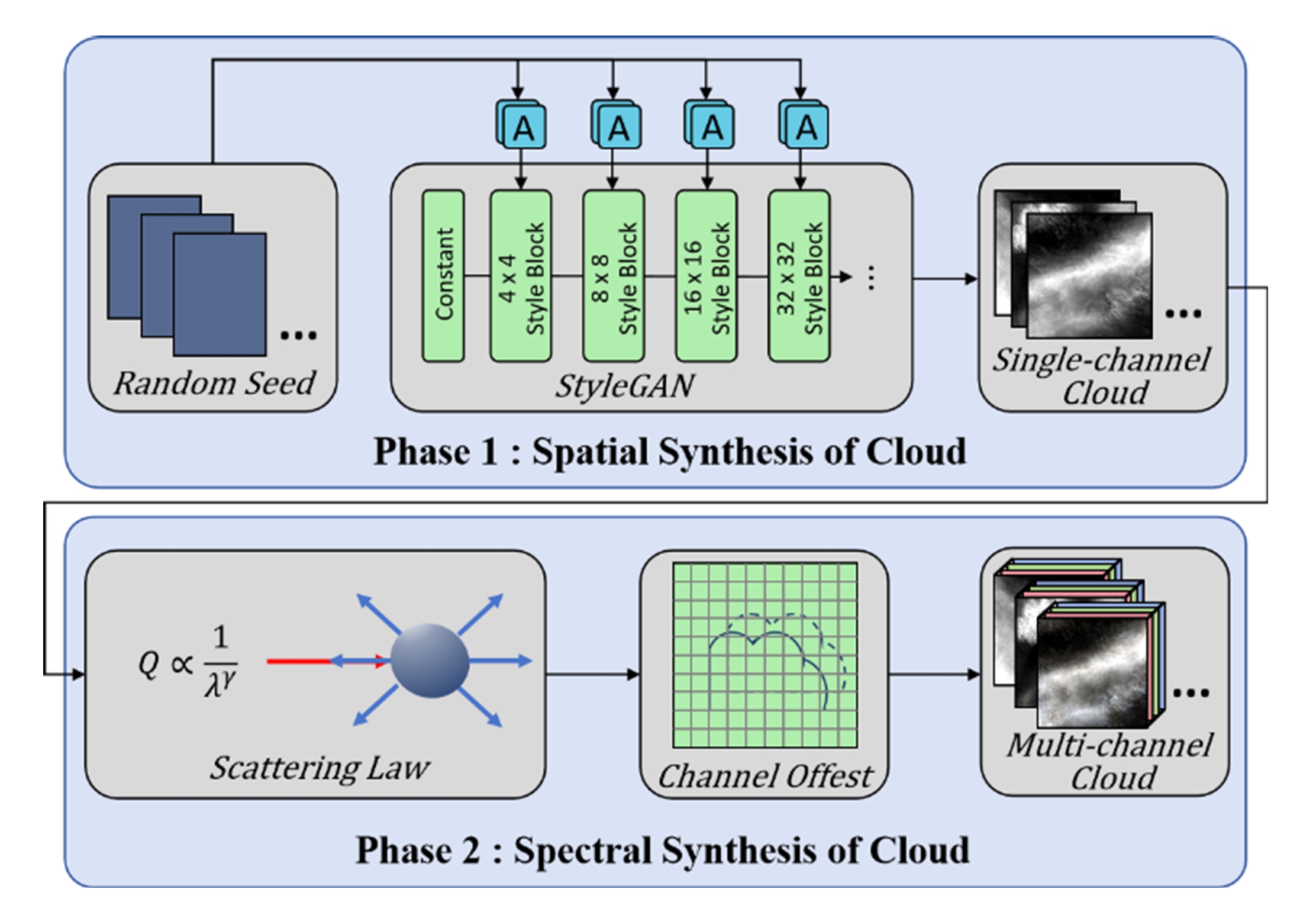}
\caption{\label{fig:fig1}Framework of the proposed PGCS.}
\end{figure}

\noindent\hspace{2em} By combining the strengths of StyleGAN in generating diverse images with the advantages of the scattering model in capturing physical properties, a hybrid method is proposed in this paper to generate infinite high-fidelity cloud images. This hybrid approach, named PGCS, involves a two-step process, as shown in Fig. \ref{fig:fig1}.

\subsection{Spatial synthesis}
\label{sec:3.1}
\setlength{\parindent}{2em}
\noindent\hspace{2em} An appropriate training dataset is crucial in enabling StyleGAN to generate target images with specific characteristics. In this context, the cirrus cloud band of the Landsat-8/9 Operational Land Imager (OLI) is specifically designed to capture the reflected light from clouds. With a wavelength range of 1.36 to 1.38 µm, the cirrus band can effectively identify and distinguish clouds from other image features. Leveraging the accurate reflection of spatial distribution characteristics by the cirrus band, we constructed a dataset of cirrus band images to train StyleGAN to generate cloud images. In this step, the training result is a cloud spatial synthesis model that can generate an infinite amount of spatially varying single-channel cloud data for the later step.

\begin{figure}
\centering
\includegraphics[width=1\linewidth]{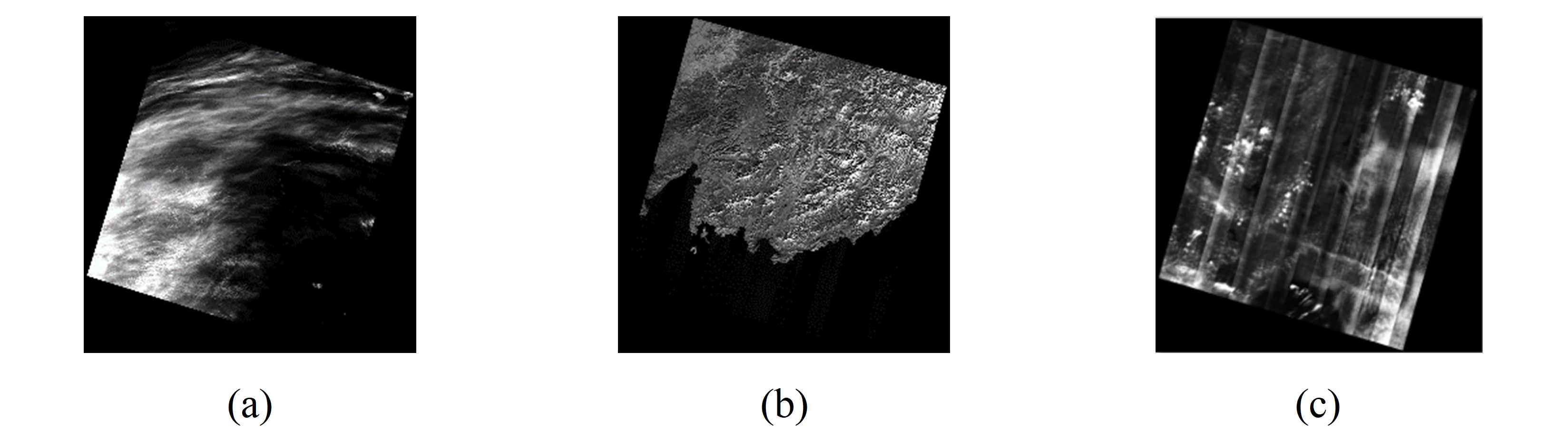}
\caption{\label{fig:fig2}Patterns of cirrus band images. (a) Pure clouds. (b) With surface information. (c) With stripe noise.}
\end{figure}

The dataset construction process for StyleGAN training involved four main steps: 1) A selection was made from a pool of approximately 20,000 Landsat-8/9 images, resulting in 170 cirrus band images that accurately captured the distribution characteristics of clouds, excluding the influence of surface information and stripe noise, as shown in Fig. \ref{fig:fig2}. 2) The selected images were radiometrically calibrated to convert digital number values to top-of-atmosphere radiation. This radiometric calibration, which utilized pertinent metadata such as sun elevation, ensured the accuracy and consistency of the dataset. This process enhanced the suitability of the dataset for precise analysis in the realm of remote sensing, as it corrected for sensor-specific biases and allowed for comparability across different images and sensors. 3) The images were cropped into patches of 521 $\times$ 521 pixels in size with a stride of 128 pixels. The adoption of a 128-pixel stride was motivated by the aim to enhance the similarity within the dataset, thereby facilitating the learning process of StyleGAN and enabling the generation of cloud images with intricate details. 4) Data cleaning was carried out by eliminating patches with a maximum value below 0.015, ensuring the exclusion of patches that predominantly represented cloud-free regions, which are irrelevant to the training of cloud generation and have the potential to introduce misleading information. As a result, approximately 18,000 patches were retained in the dataset.

Due to the distinct value ranges between remote sensing images and natural images, a customized normalization approach was adopted for the cloud spatial synthesis dataset. Statistical analysis showed that the radiation values of clouds in the cirrus cloud band are primarily between 0 and 0.1. Therefore, the pixel values were normalized by dividing by 0.05 and subtracting 1, resulting in a range of -1 to 1. This normalization method effectively captured the unique characteristics of clouds while maintaining compatibility with the StyleGAN framework. By preserving the relative distribution of values, this approach facilitated the generation of realistic and diverse cloud images. Thickness adjustment was achieved by applying threshold limits to the output single-channel clouds images, while linear stretching of the output pixel values was also employed to adjust the thickness.

\subsection{Spectral synthesis}
\label{sec:3.2}
\setlength{\parindent}{2em}
\noindent\hspace{2em} According to Section \ref{sec:2.1}, Chavez Jr \cite{RN4} demonstrated a negative correlation between the value of  parameter $\gamma$ and the radiation of clouds. Zhang, et al. \cite{RN31}  proposed a pixel-wise technique for determining the value of parameter $\gamma$. Based on the above two works, a function $\gamma=F(C_r)$ depicting the relationship between parameter $\gamma$ and the reference cloud radiation $C_r$ can be fitted to realize the spectral synthesis of clouds.

Specifically, this function can be substituted into Equation \ref{eq:eq2} to calculate the target cloud radiation $C_t=(\frac{\lambda_r}{\lambda_t})^F(C_r) \cdot C_r$ by changing the target wavelength $\lambda_t$. The reference wavelength $\lambda_r$ was 1.375 µm in this study as the reference cloud radiation $C_r$ was generated by the synthesis model trained using cirrus band images in the first phase.

\begin{figure}
\centering
\includegraphics[width=1\linewidth]{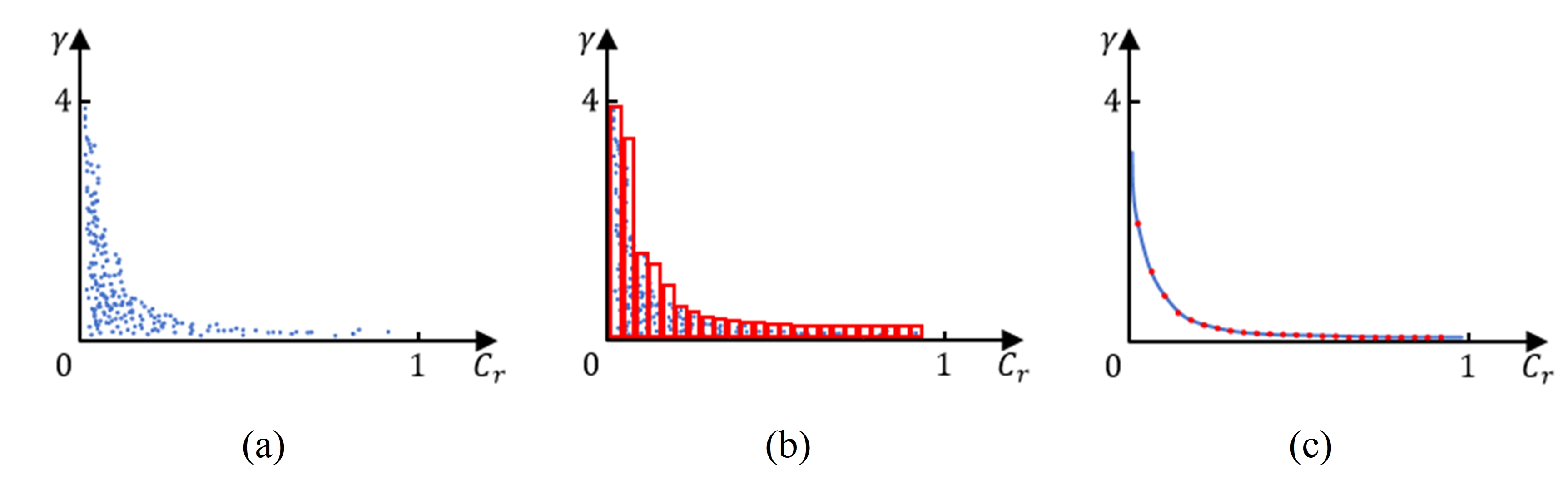}
\caption{\label{fig:fig3}Schematic representation of LSGF.}
\end{figure}

Therefore, the current task is to fit the function $\gamma=F(C_r)$ . The cirrus band images and the discrete pixel-wise parameter $\gamma$ values calculated by Zhang’s method provide the necessary samples for fitting the function. Fig. \ref{fig:fig3}a shows a schematic scatter diagram drawn based on the calculation results of Zhang, et al. \cite{RN31} . The scatter distribution between parameter $\gamma$ and the reference cloud radiation $C_r$ is unevenly distributed, with a notable concentration in the low radiation range. This is due to the fact that the radiation values of cirrus clouds are inherently unevenly distributed, often clustering in regions of lower intensity. To avoid the influence of an uneven distribution on the statistical results, a local statistics and global fitting method (LSGF) is proposed that optimally captures both local and global features to provide a more representative function.

LSGF divides the reference cloud radiation $C_r$ into several equal subsets, as shown in Fig. \ref{fig:fig3}b. The statistical measures of parameter $\gamma$ and reference cloud radiation $C_r$ such as the mode, median, and mean, are then calculated, respectively, for each subset. Finally, the aggregated statistics are used to fit a global model, as shown in Fig. \ref{fig:fig3}c. LSGF offers a robust solution for fitting models to datasets with unevenly distributed points. By focusing on statistical aggregates within subsets and applying a global fitting approach, LSGF mitigates the bias introduced by high-density regions, providing a more accurate representation of the underlying relationships of the data.

\begin{figure}
\centering
\includegraphics[width=1\linewidth]{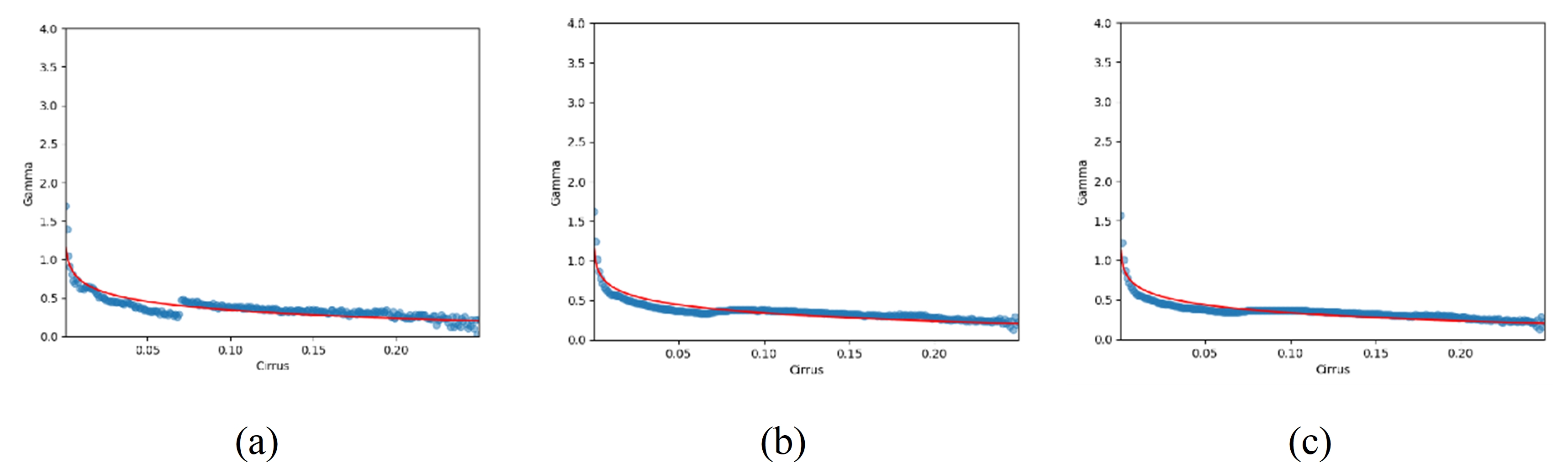}
\caption{\label{fig:fig4}Scatter plots of the statistical measures. (a) Mode. (b) Median. (c) Mean.}
\end{figure}

For practical implementation, LSGF was applied to analyze the relationship in approximately 93.8 million pixels obtained from real cloudy images. As a result, three sets of statistical measures were derived, as illustrated in Fig. \ref{fig:fig4}, with each set comprising approximately 250 paired statistical measures. Specifically, the three sets of measures corresponded to the mode, median, and mean of parameter $\gamma$. The mean of the reference cloud radiation $C_r$ was used, due to the minimal variation within each subset.

The experimental findings indicate the equivalence of the fitting expressions for the mode, median, and mean, as depicted in Equation \ref{eq:eq3}. The accuracy of the three sets of fittings is quantified by the respective $R^2$ values of 0.78, 0.80, and 0.81. Additional verification of the fitting expressions was conducted, as described in Section \ref{sec:4.2}.

\begin{equation}
\gamma = -0.14\cdot\ln(C_r)\label{eq:eq3}
\end{equation}

Given the above, Equation \ref{eq:eq2} can be expressed as shown in Equation \ref{eq:eq4}, which serves the purpose of expanding single-channel images to multi-channel cloud images.

\begin{equation}
C_t = (\frac{\lambda_r}{\lambda_t})^{-0.14\cdot\ln(C_r)} \cdot C_r\label{eq:eq4}
\end{equation}

\noindent where $C_t$ represents the cloud radiation of the target band, $\lambda_t$ represents the wavelength of the target band, and $\lambda_r$ takes the value of1.375µm.

Moreover, the parallax issue occurs with many satellite sensors because the different spectral bands are configured within the focal plane module. Each band observes the same ground feature from a different angle, causing discrepancies \cite{RN85}. The parallax error was simulated through channel offset after the multi-channel cloud radiation was obtained. By analyzing the offsets in real cloud data, the maximum channel offset value was first determined \cite{RN83}. The actual offset for each channel was then randomly sampled within this maximum limit.

\subsection{Paired data synthesis}
\label{sec:3.3}
\setlength{\parindent}{2em}
\noindent\hspace{2em} Synthesized clouds have a data augmentation effect, which can be used for degraded data simulation to enhance the capabilities of algorithms for tasks such as classification, recognition, and segmentation. Moreover, many cloud related tasks in remote sensing images are driven by paired data, such as cloud detection, cloud correction and cloud removal. On the basis of cloud synthesis, paired data, including clear ground, clouds, and cloudy images, can be generated.

The observed image $\rho$ with cloud cover can be expressed as $\rho=G+C$, where $G$ is the ground surface radiation and $C$ is the cloud radiation. The surface radiation can be collected from clear remote sensing images and the cloud radiation can be synthesized by PGCS. A cloudy image can then be synthesized by adding any cloud and surface image together. In addition, data augmentation techniques such as rotation, horizontal flipping, and vertical flipping can be applied to the synthesized cloudy images. These transformations effectively augment the data, resulting in a more comprehensive and representative dataset for subsequent tasks. The degradation degree of cloudy images can be controlled by adjusting the thickness of the single-channel cloud images.

In summary, PGCS allows for the generation of clouds with different spatial patterns by changing the random input numbers of the spatial synthesis model. Moreover, numerical statistical methods can be used to integrate the existing physical mechanism laws, to ensure the spectral authenticity of the synthetic data. In contrast to the synthesis methods based on cloud retrieval, PGCS allows for the creation of diverse clouds using a single set of parameters, thus augmenting the training dataset and enhancing its diversity. This significantly reduces the reliance on large pre-existing training datasets, which often pose challenges related to data storage and management. In addition, the efficiency of cloud synthesis is markedly improved by minimizing the need to download and preprocess cloud images \cite{RN88}.

\section{Results}
\label{sec:4}
\noindent\hspace{2em} The synthetic results of PGCS are provided in Section \ref{sec:4.1}. Three sets of experiments were conducted to thoroughly evaluate the PGCS method. The accuracies in the spatial synthesis phase and the spectral synthesis phase are evaluated in Section \ref{sec:4.2}. In order to evaluate the synthesis quality, PGCS is compared to three other cloud synthesis methods in Section \ref{sec:4.3}. Finally, two cloud correction methods extended from PGCS, i.e., PGCS$_\text{M}$ and PGCS$_\text{D}$, are compared to three other cloud correction methods in Section \ref{sec:4.4}. 

\subsection{Cloud synthesis}
\label{sec:4.1}
\begin{figure}
\centering
\includegraphics[width=1\linewidth]{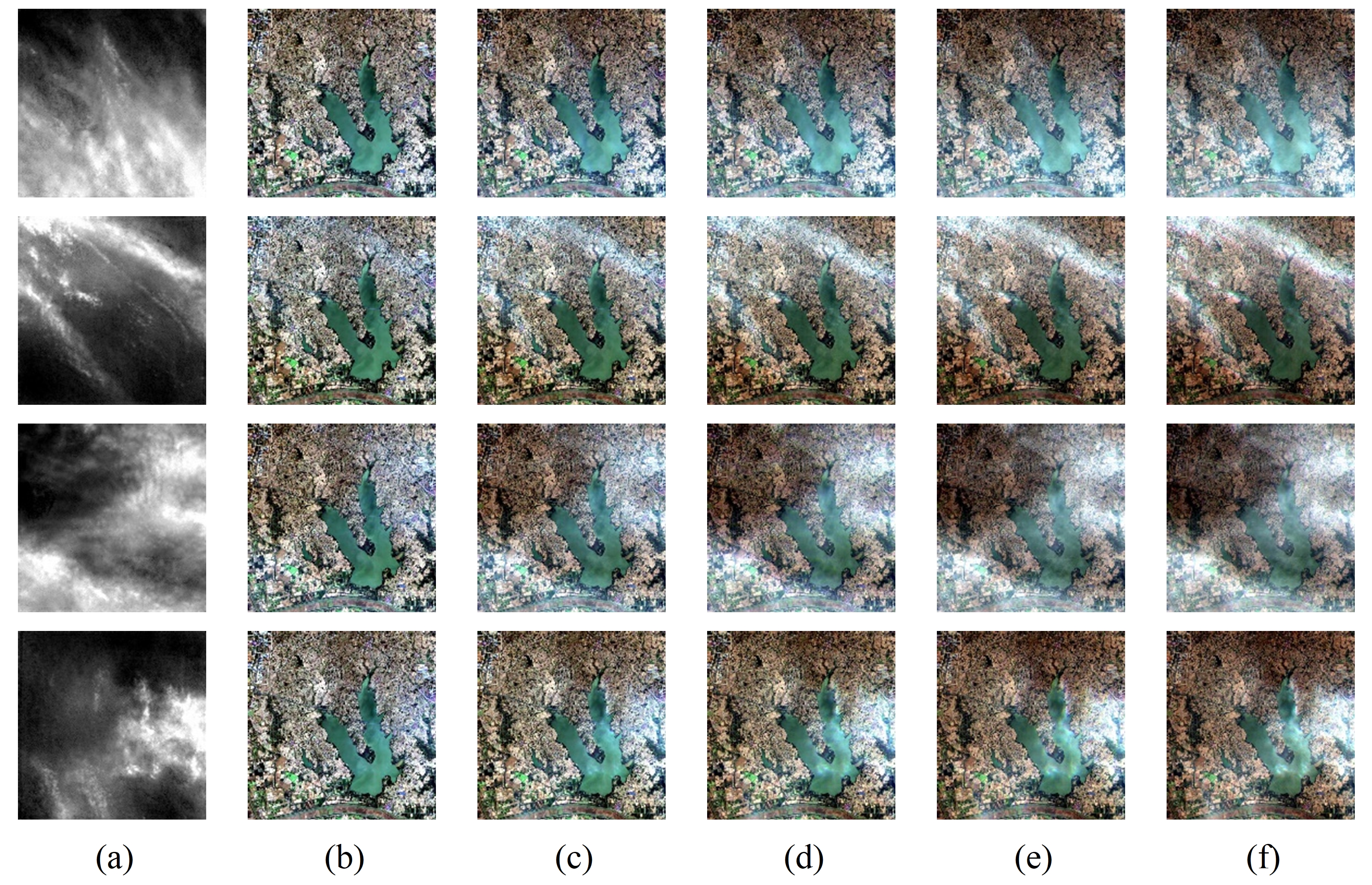}
\caption{\label{fig:fig5}Various synthetic results generated by PGCS. (a) Cloud images with different shapes. (b)-(d) Cloudy images with different cloud thicknesses.}
\end{figure}

\setlength{\parindent}{2em}
\noindent\hspace{2em} The cloud spatial synthesis model can generate single-channel cloud images exhibiting a diverse array of random shapes as shown in Fig. \ref{fig:fig5}a. The same clear surface land image was selected as the background to better display the synthesized cloudy images with diverse thicknesses, as shown in Fig. \ref{fig:fig5}b-f.

\subsection{Verification of synthetic cloud}
\label{sec:4.2}
\setlength{\parindent}{2em}
\noindent\hspace{2em} The verification of the synthetic cloud generated by PGCS encompassed two main parts: spatial fidelity and spectral precision. Due to the cirrus band carried on Landsat-8/9, the real radiation data of cirrus clouds can be obtained. Therefore, in this section we describe how we employed cirrus clouds to quantitatively validate the synthetic accuracy of PGCS.

\begin{figure}
\centering
\includegraphics[width=1\linewidth]{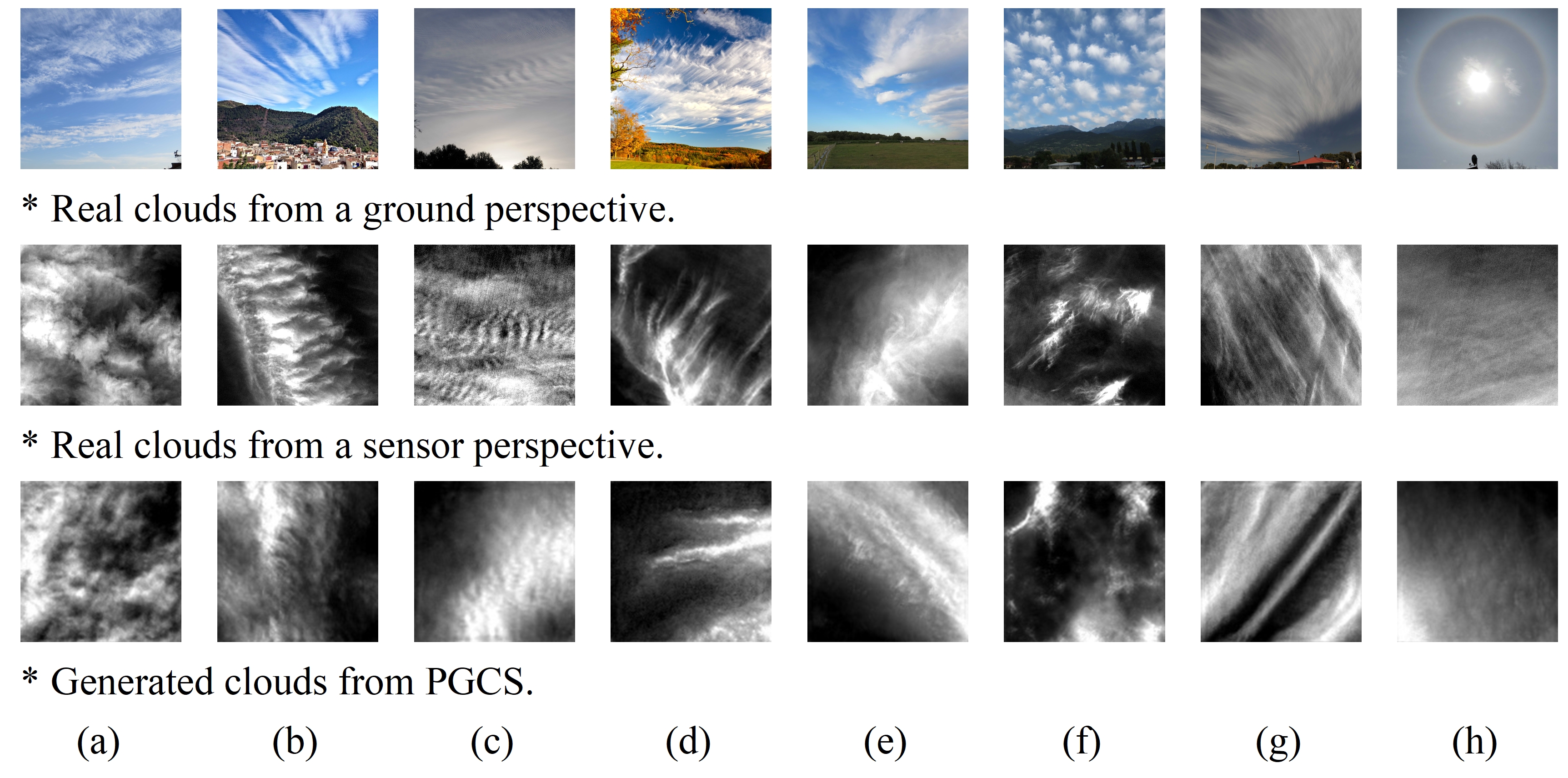}
\caption{\label{fig:fig6}Cloud with various shapes from different perspectives. (a) Castellanus. (b) Lenticularis. (c) Stratiformis. (d) Uncinate. (e) Spissatus. (f) Flocculent. (g) Fibrous. (h) Nebulous.}
\end{figure}

To demonstrate the ability of the cloud spatial synthesis model to generate clouds with realistic and diverse shapes, Fig. \ref{fig:fig6} presents a comparison between the generated clouds and real clouds. The first row shows real clouds from a ground perspective, categorized based on the guidelines provided by the World Meteorological Organization’s International Cloud Atlas (\url{https://cloudatlas.wmo.int/en/descriptions-of-clouds.html}). The second row presents the appearances of real clouds from a sensor perspective, which were captured by the cirrus band. The third row displays the clouds synthesized by PGCS, which were selected from the generated dataset. The comparison shows that the cloud spatial synthesis model successfully captures the different types of clouds and shows high fidelity in simulating different spatial morphologies.

In addition, histograms were plotted to visualize the data distribution between the real clouds and the generated clouds. The overlap rate of the two histograms was calculated to quantitatively assesses the accuracy of the spatial morphology simulation \cite{RN79}. A higher overlap rate indicates greater similarity between the data, thereby signifying superior synthesis results.

\begin{figure}
\centering
\includegraphics[width=1\linewidth]{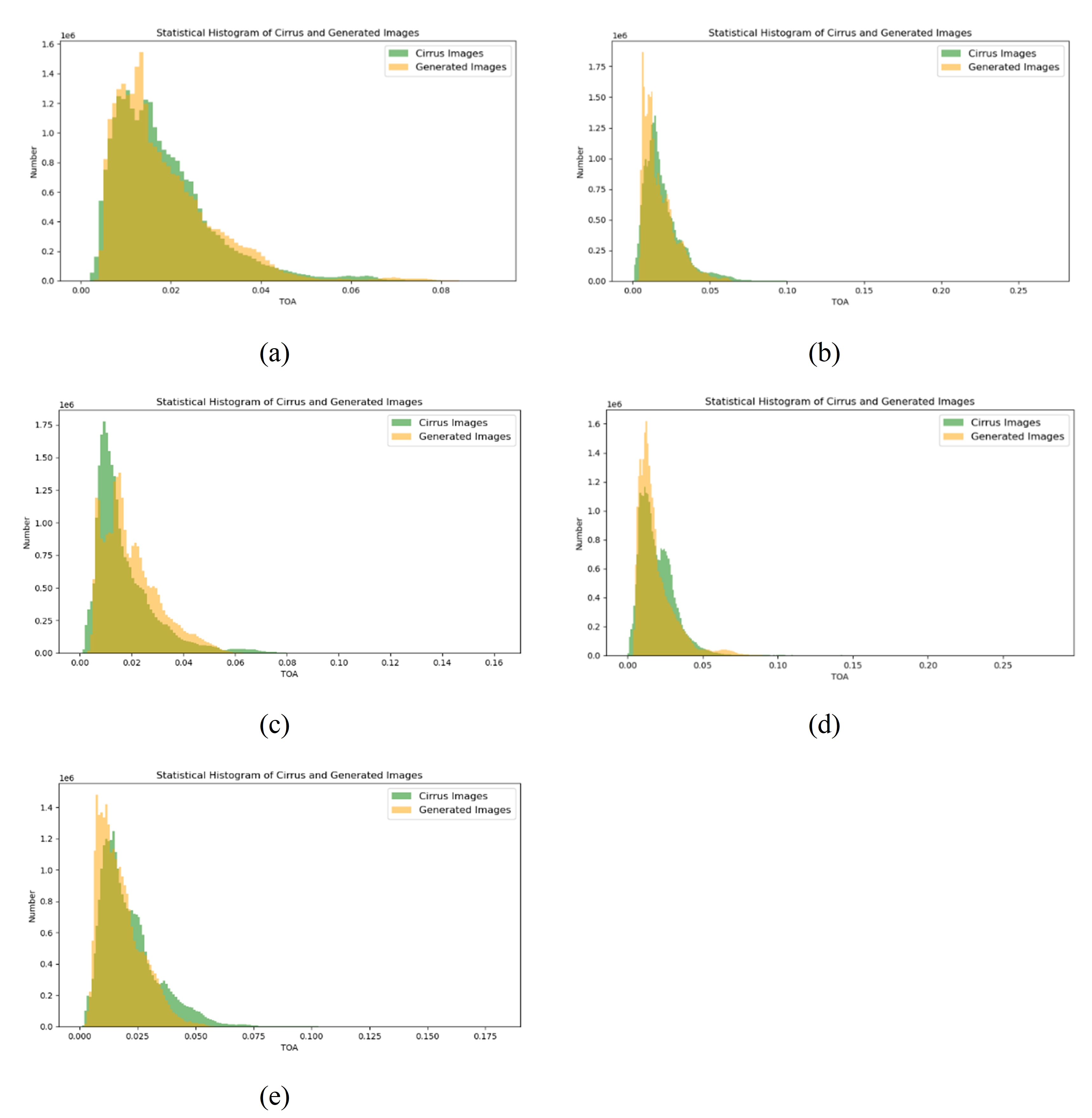}
\caption{\label{fig:fig7}Statistical histograms of cirrus clouds and generated images. (a) Experiment 1 with an overlap rate of 91.89$\%$. (b) Experiment 2 with an overlap rate of 84.00$\%$. (c) Experiment 3 with an overlap rate of 80.06$\%$. (d) Experiment 4 with an overlap rate of 84.73$\%$. (e) Experiment 5 with an overlap rate of 85.65$\%$.}
\end{figure}

The histograms of the real cirrus clouds and the generated clouds were superimposed on the same graph, sharing a common coordinate system, as illustrated in Fig. \ref{fig:fig7}. Specifically, the green histogram represents the distribution of 100 real cloud images randomly selected from the dataset described in Section \ref{sec:3.1}, which contains 18,000 real cirrus cloud images. Meanwhile the yellow histogram represents the distribution of 100 cloud images generated by the cloud spatial synthesis model using random seeds. The overlap rate of the two histograms was determined by calculating the ratio of the overlapping area to the total area of the real cirrus cloud histogram.

In order to ensure robustness and minimize chance occurrences, a total of five experimental replicates were conducted. For each experiment, the rate consistently exceeded 80$\%$. Moreover, with a mean overlap rate of 85.27$\%$, the significant concordance between the simulated and actual datasets was verified. This outcome further underscores the scientific validity of the spatial synthesis phase of PGCS.

In the validation of the spectral precision, the focus lies on simulating the radiation of cloud in multiple spectral bands. The proposed spectral synthesis method can directly estimate corresponding multi-channel cloud images from the cirrus cloud images. Moreover, accurate estimation of the real multi-channel cloud radiation was achieved by conducting pixel-to-pixel calculations in a previous study \cite{RN31}. Given that true multi-channel cloud data are unavailable, the cloud estimates generated by Zhang’s method can be used as the reference for evaluating the performance of the proposed approach. The root mean square error (RMSE) metric is utilized here to evaluate the similarity between the synthesized and reference multi-channel cloud images. A smaller RMSE value indicates a higher level of consistency and suggests a superior performance of the synthesis method.

\begin{figure}
\centering
\includegraphics[width=1\linewidth]{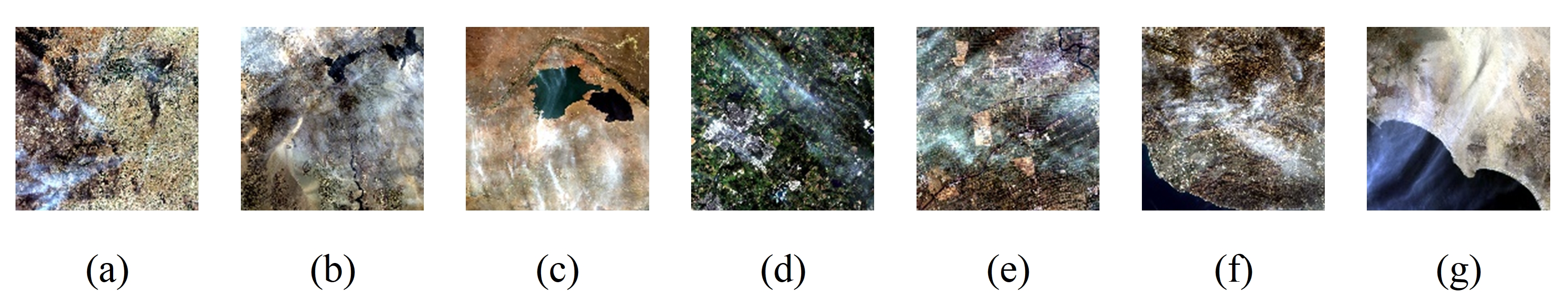}
\caption{\label{fig:fig8}True-color composite cloudy images for seven different scenes. (a)-(g) Scenes1-7.}
\end{figure}

\begin{figure}
\centering
\includegraphics[width=1\linewidth]{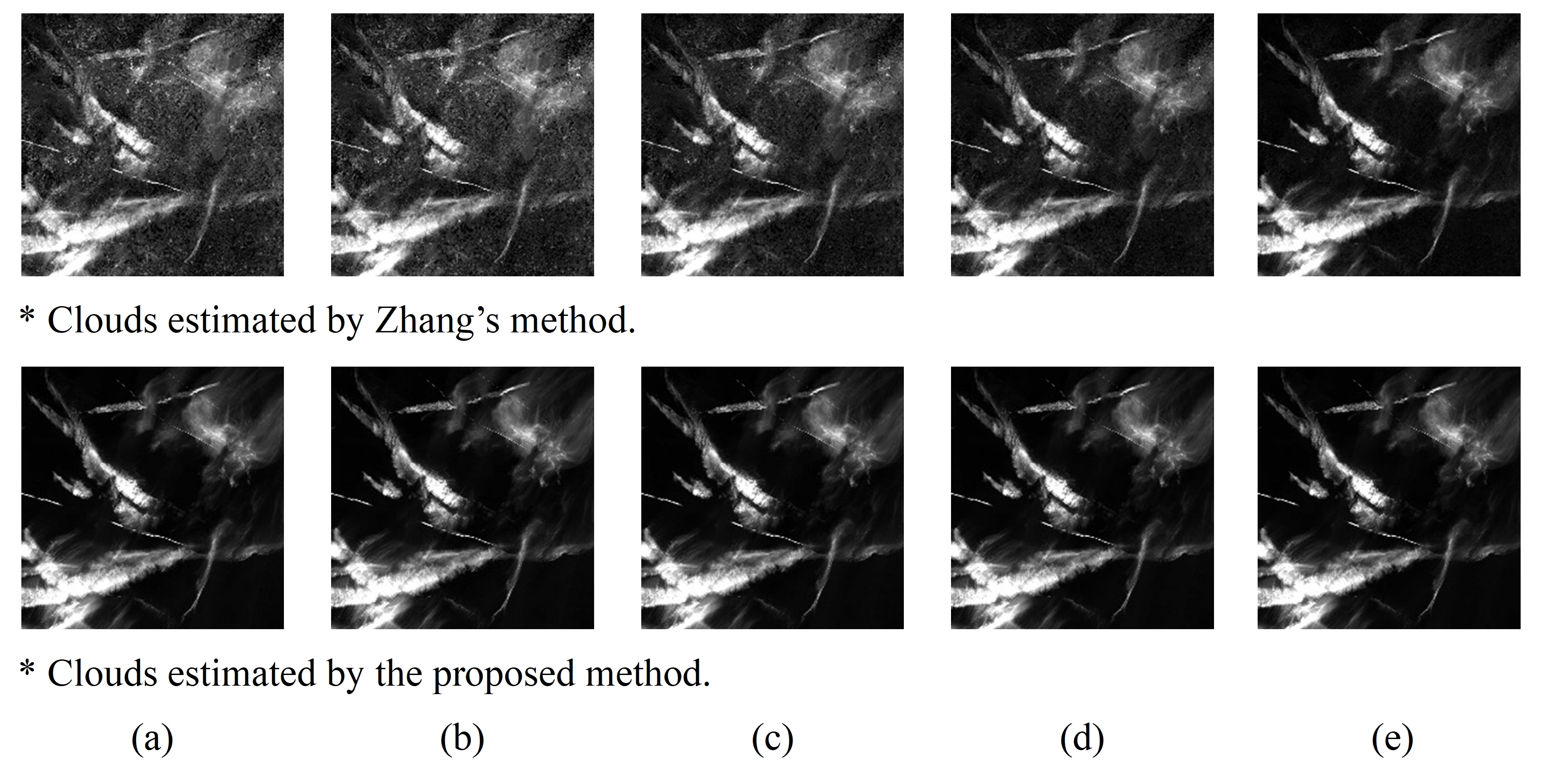}
\caption{\label{fig:fig9}Estimated clouds in multi-spectral bands of an example scene. (a)-(e) Coastal, blue, green, red, and near-infrared bands, respectively.}
\end{figure}

The experiment was conducted across seven cloudy images with distinct land cover scenes, as illustrated in Fig. \ref{fig:fig8}. To provide a clearer comparison of the two cloud estimation results, Fig. \ref{fig:fig9} presents the band-by-band cloud images for Scene 1 from Fig. \ref{fig:fig8}. The proposed method estimates clouds directly from the cirrus band, ensuring that the results are free from any mixed surface information. While the results of the proposed method appear visually similar across the different bands, they exhibit substantial numerical differences. Detailed assessments of the individual bands and the overall performance in these seven scenes are summarized in Table \ref{tab:table1}. Notably, all the results exhibit RMSE values that are close to zero, which confirms the exceptional accuracy achieved in the spectral synthesis process.

\begin{table}
    \centering
\caption{\label{tab:table1}Evaluation of the RMSE values for the cloud spectral synthesis.}
    \begin{tabular}{>{\centering\arraybackslash}p{1.5cm}>{\centering\arraybackslash}p{1.5cm}>{\centering\arraybackslash}p{1.5cm}>{\centering\arraybackslash}p{1.5cm}>{\centering\arraybackslash}p{1.5cm}>{\centering\arraybackslash}p{1.5cm}>{\centering\arraybackslash}p{1.5cm}}
    \toprule
         &  Overall&  Band 1&  Band 2&  Band 3&  Band 4& Band 5\\
         \midrule
         Scene 1&  0.007&  0.010&  0.009&  0.006&  0.004& 0.002\\ 
         Scene 2&  0.012&  0.015&  0.015&  0.011&  0.009& 0.005\\ 
         Scene 3&  0.012&  0.015&  0.015&  0.012&  0.009& 0.005\\ 
         Scene 4&  0.006&  0.008&  0.008&  0.006&  0.005& 0.003\\ 
         Scene 5&  0.011&  0.015&  0.014&  0.011&  0.009& 0.005\\  
         Scene 6&  0.013&  0.016&  0.016&  0.012&  0.010& 0.006\\ 
         Scene 7&  0.014&  0.018&  0.017&  0.013&  0.011& 0.006\\
         \bottomrule
         \multicolumn{7}{l}{*Bands 1-5 represent coastal, blue, green, red, and near-infrared bands respectively.}\\    
    \end{tabular}
    
\end{table}

In additionally, it should be mentioned that Equation \ref{eq:eq4} from the spectral synthesis phase can be used not only for cloud synthesis but also for thin cloud correction when the cirrus band is available. By subtracting the estimated cloud radiation from the cloudy images, clear surface information can be obtained. The detailed effects of cloud correction are discussed in depth in Section \ref{sec:4.4}.

\subsection{Cloud synthesis comparison}
\label{sec:4.3}
\setlength{\parindent}{2em}
\noindent\hspace{2em} The proposed method was compared with three representative methods to assess their performance in cloud synthesis, i.e., a computer generation method \cite{RN61}, a transmission estimation method \cite{RN62}, and a cloud self-subtraction method \cite{RN63}. Due to the distinct radiation characteristics present in the bands of both the clouds and the surface, thin cloud synthesis is more challenging than thick cloud synthesis. The thin cloud correction task is used to compare the effects of the cloud synthesis in this section, which includes a visual analysis of the synthetic training datasets and a qualitative and quantitative evaluation of the thin cloud correction results for the real test datasets.

Each of the four synthesis methods was used to produce four synthetic training datasets containing an equal number of “cloudy $\&$ cloud-free” image pairs. All the cloud-free patches in these training datasets were derived from real clear images with dimensions of 512 $\times$ 512 pixels. These patches encompassed a variety of surface features, including urban areas, forests, bare soil, and water bodies. The training cloudy patches were generated by synthesized clouds and real cloud-free patches. These training datasets were then used to train four thin cloud correction models based on a residual network (ResNet) \cite{RN47}. 

The four models were then utilized to correct the thin clouds in the real testdatasets, which consisted of real cloudy images and their corresponding time-phase cloud-free images. The cloudy images were used as input for the correction models, while the cloud-free images served as reference data to evaluate the correction performance of the models based on the differences between the outputs of the models and the cloud-free images. These diverse test images present a wide range of surfaces under varying atmospheric conditions, posing a challenging task for supervised deep-learning algorithms. The better the thin cloud correction ability of the model, the higher the similarity between the synthetic training cloudy image and the real test cloudy image, indicating a superior synthesis ability. 

\begin{figure}
\centering
\includegraphics[width=1\linewidth]{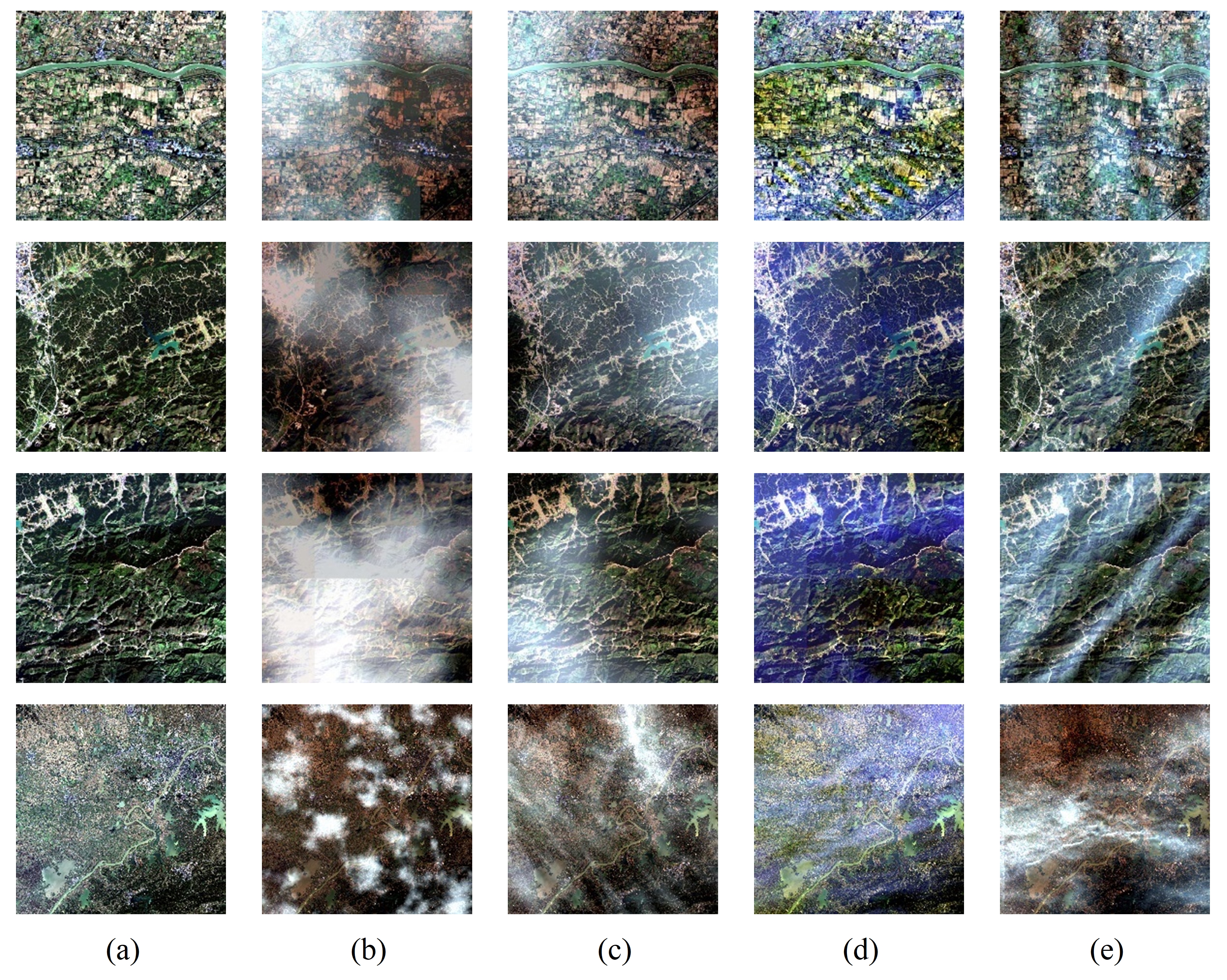}
\caption{\label{fig:fig10}Examples of synthetic cloudy images. (a) Cloud-free image. (b) Cloudy image synthesized by the computer generation method. (c) Cloudy image synthesized by the transmission estimation method. (d) Cloudy image synthesized by the cloud self-subtraction method. (e) Cloudy image synthesized by PGCS.}
\end{figure}

Fig. \ref{fig:fig10} illustrate the results of the different synthesis methods to provide an intuitive comparison. The first to third rows of Fig. \ref{fig:fig10} present images of size 512 $\times$ 512 pixels, while the fourth row shows images of size 5120 $\times$ 5120 pixels to more clearly highlight the differences between the methods. Based on the observation of Fig. \ref{fig:fig10}, it can be concluded that the clouds synthesized by the computer generation method appear unrealistic and do not match the actual cloud distribution patterns. In contrast, the clouds produced by the transmission estimation method are excessively smooth and lack detailed textures, due to the application of filtering during the estimation of the transmission map. Similarly, the cloud self-subtraction method also yields overly smooth results. Furthermore, the clouds generated by the cloud self-subtraction method exhibit a blue tint, indicating spectral distortion.

\begin{figure}
\centering
\includegraphics[width=1\linewidth]{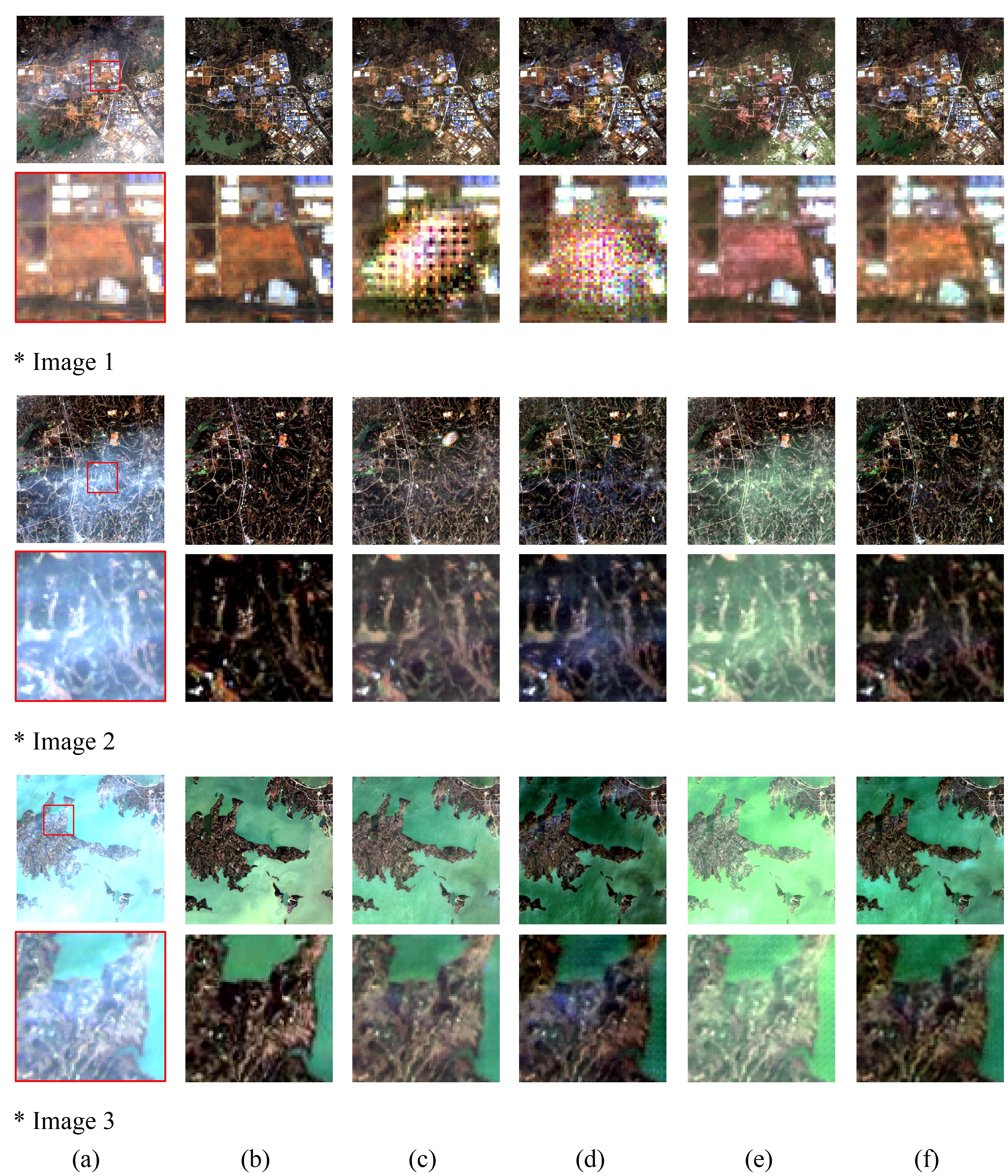}
\caption{\label{fig:fig11}Thin cloud correction results based on the various synthesis methods. (a) Real cloudy images. (b) Real clear image. (c) Computer generation method. (d) Transmission estimation method. (e) Cloud self-subtraction method. (f) PGCS.}
\end{figure}

Fig. \ref{fig:fig11} presents three representative thin cloud-contaminated images to assess the performance of the different methods, with the level of cloud degradation progressively increasing from the first image to the third image. The cloud self-subtraction method fails to effectively remove the clouds and only changes the color of the original image, as illustrated in the fifth column. In the results for the second and third images, the other three methods have completely corrected the clouds. However, the results from the computer generation method and the transmission estimation method exhibit noticeable distortions in the high-brightness areas. In the results for the second and third images, the computer generation method shows relatively good cloud removal effectiveness and global consistency despite minor color distortions. The transmission estimation method shows incomplete cloud removal in dealing with the uneven cloud coverage. For PGCS, all the cloud correction results demonstrate superior cloud removal completion, ground fidelity, and global consistency. This suggests that PGCS is capable of effectively adapting to complex thin cloud contamination in images. The quantitative evaluation metrics employed here for this experiment include the peak signal-to-noise ratio (PSNR), structural similarity index (SSIM), correlation coefficient (CC), spectral angle mapper (SAM), and RMSE. PGCS exhibits the best performance in the test dataset across all the metrics, as shown in Table \ref{tab:table2}.

\begin{table}
    \centering
\caption{\label{tab:table2}Evaluation of the different synthesis methods.}
    \begin{tabular}{>{\centering\arraybackslash}p{3.5cm}>{\centering\arraybackslash}p{1.5cm}>{\centering\arraybackslash}p{1.5cm}>{\centering\arraybackslash}p{1.5cm}>{\centering\arraybackslash}p{1.5cm}>{\centering\arraybackslash}p{1.5cm}}
    \toprule
         &  PSNR↑&  SSIM↑&  CC↑&  SAM↓& RMSE↓\\
         \midrule
         Computer generation&  \underline{35.6751}&  \underline{0.8713}&  \underline{0.7794}&  \underline{2.8585}& \underline{0.0165}\\
         Transmission estimation&  34.7138&  0.8664&  0.7753&  3.5066& 0.0184\\
         Cloud self-subtraction&  31.2437&  0.7747&  0.6352&  3.6184& 0.0274\\
         PGCS&  \textbf{36.3603}&  \textbf{0.8819}&  \textbf{0.8214}&  \textbf{2.5899}& \textbf{0.0152}\\
         \bottomrule
         \multicolumn{6}{l}{*Best results are in \textbf{bold}, second-best results are \underline{underlined}.}\\ 
    \end{tabular}
\end{table}

The superiority of PGCS can be attributed to two primary factors. Firstly, PGCS utilizes real observed cloud distributions as guidance, ensuring that the generation process accurately captures the authenticity of cloud formations. In contrast, the computer generation method lacks any guiding information. Meanwhile, both the transmission estimation method and the cloud self-subtraction method rely on the cloud reference derived from cloud retrieval calculations. However, these calculations are prone to errors, which can introduce inaccuracies into the synthesis results. Furthermore, certain algorithmic steps, such as filtering procedures, may inadvertently blur the spatial distribution of the clouds, further compromising the accuracy of synthetic cloud generation. The PGCS method incorporates a diverse range of cloud features during the generation process, effectively capturing the complexity and variability of cloud formations. Notably, PGCS possesses the unique ability to generate an infinite variety of clouds. This integration significantly enhances the generalization capabilities of the trained model, thereby improving its robustness and efficacy in addressing a diverse spectrum of cloud correction challenges.

\subsection{Application in cloud correction}
\label{sec:4.4}
\setlength{\parindent}{2em}
\noindent\hspace{2em} Thin cloud correction is a crucial step in the processing of remote sensing images. It is therefore essential to evaluate the effectiveness of the proposed cloud synthesis method in supporting cloud correction. As described above, with the support of the cirrus band, the cloud correction results can be obtained directly by estimating parameter  through Equation \ref{eq:eq4}. In addition, deep learning methods can achieve thin cloud correction using synthesized datasets, thereby reducing the reliance on the cirrus band. These two methods extended from PGCS are named PGCS$\_$M (model driven) and PGCS$\_$D (data driven). 

In order to demonstrate that PGCS can advance the development of the subsequent task, PGCS$\_$M and PGCS$\_$D were compared to three advanced thin cloud correction methods. The CR-NAPCT method use a noise-adjusted principal components transform model to remove thin cloud in the cloudy images \cite{RN80}. The DBOASM method enhances image visibility and partially corrects spectral bias by employing an optimized atmospheric scattering model for effective dehazing \cite{RN81}. The GTCCM method is a general thin cloud correction method for satellite VNIR images that combines statistical information with a scattering model to accurately estimate and remove cloud contamination \cite{RN82}. 

\begin{figure}
\centering
\includegraphics[width=1\linewidth]{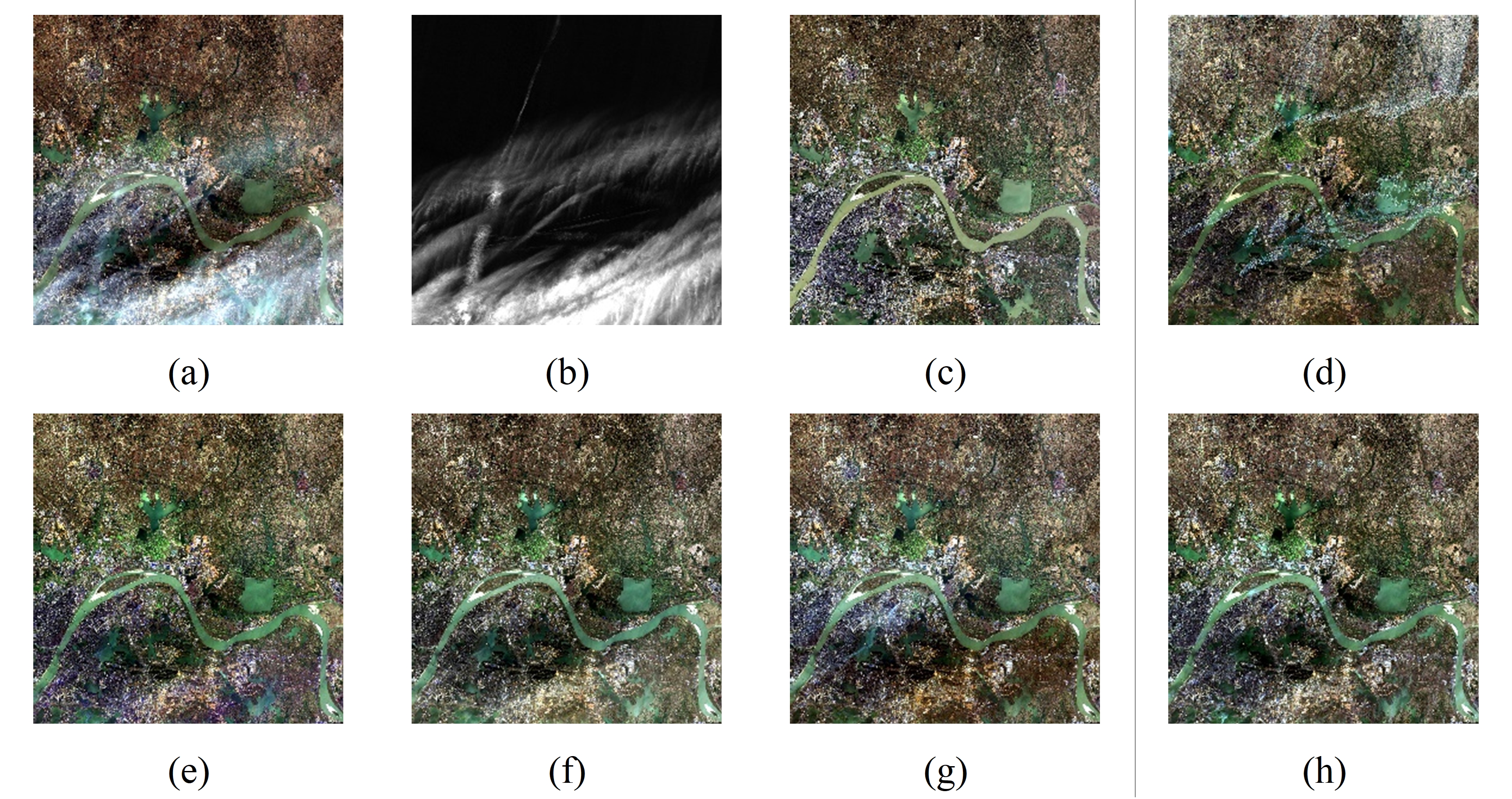}
\caption{\label{fig:fig12}Cirrus cloud correction results compared with various traditional methods. (a) Cloudy image on November 26, 2021. (b) The cirrus band. (c) Reference clear image on November 10, 2021. (d) CR-NAPCT. (e) DBOASM. (f) GTCCM. (g) PGCS$\_$M. (h) PGCS$\_$D.}
\end{figure}

\begin{figure}
\centering
\includegraphics[width=1\linewidth]{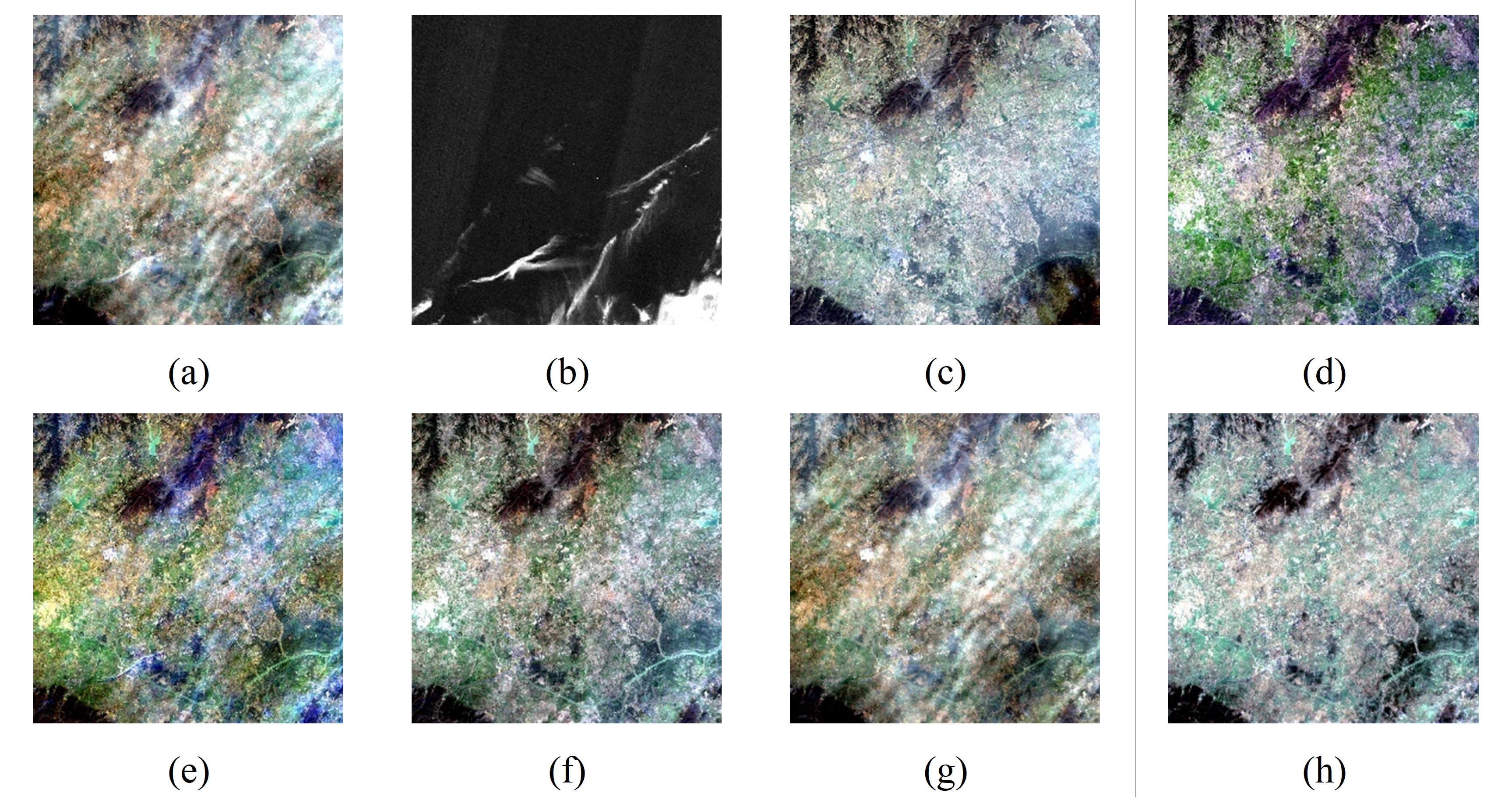}
\caption{\label{fig:fig13}Thin cloud correction results compared with the various traditional methods. (a) Cloudy image on February 19, 2023. (b) The cirrus band. (c) Reference clear image on February 3, 2023. (d) CR-NAPCT. (e) DBOASM. (f) GTCCM. (g) PGCS$\_$M. (h) PGCS$\_$D.}
\end{figure}

Two real Landsat-8/9 OLI images with diverse cloud and land surfaces were chosen for the experiments. The region of Fig. \ref{fig:fig12}a is covered by cirrus clouds, which can be detected by the cirrus band, as shown in Fig. \ref{fig:fig12}b. The thin cloud of Fig. \ref{fig:fig13}a is distinct from the cirrus clouds and cannot be identified by the cirrus band, as shown in Fig. \ref{fig:fig13}b. The choice of these two images demonstrates that PGCS is adaptable to a variety of thin cloud degradation scenarios. The optimal parameters of each method were selected during the experiments. 

The cirrus band was used to create a cloud mask that guided the cloud correction process of the CR-NAPCT method. However, CR-NAPCT shows over-correction in areas with cirrus clouds, resulting in cloud artifacts in cloud-free regions, as illustrated in Fig. \ref{fig:fig12}d. In contrast, both the DBOASM and the GTCCM methods operate without cloud masks. While DBOASM exhibits slight color shifts in Fig. \ref{fig:fig12}e, GTCCM and the proposed method demonstrated a superior visual quality. The quantitative evaluation results, summarized in Table \ref{tab:table3}, indicate that the proposed PGCS$\_$D method achieves the best performance, and the proposed PGCS$\_$M method achieves the second-best performance.

\begin{table}
    \centering
\caption{\label{tab:table3}Evaluation of the different correction methods in cirrus cloud-covered region.}
    \begin{tabular}{>{\centering\arraybackslash}p{3.5cm}>{\centering\arraybackslash}p{1.5cm}>{\centering\arraybackslash}p{1.5cm}>{\centering\arraybackslash}p{1.5cm}>{\centering\arraybackslash}p{1.5cm}>{\centering\arraybackslash}p{1.5cm}}
    \toprule
         &  PSNR↑&  SSIM↑&  CC↑&  SAM↓& RMSE↓\\
         \midrule
         CR-NAPCT&  19.5854&  0.0485&  0.2891&  75.0540& 0.1049\\
         DBOASM&  30.6556&  0.6579&  0.4032&  4.4094& 0.0293\\
         GTCCM&  30.3901&  0.6714&  \underline{0.4246}&  \underline{4.1989}& 0.0302\\
         PGCS$\_$M& \underline{31.6394}& \underline{0.6754}& 0.4184& 4.2307& \underline{0.0262}\\
         PGCS$\_$D&  \textbf{32.3222}&  \textbf{0.6946}&  \textbf{0.4366}&  \textbf{4.1824}& \textbf{0.0242}\\
         \bottomrule
         \multicolumn{6}{l}{*Best results are in \textbf{bold}, second-best results are \underline{underlined}.}\\  
    \end{tabular}
\end{table}

The non-cirrus cloud correction results shown in Fig. \ref{fig:fig13} correspond to the quantitative evaluation results presented in Table \ref{tab:table4}. A mask of all the cloudy pixels was created based on actual conditions for CR-NAPCT. The visual correction results of CR-NAPCT exhibit spectral distortion, with all the values negative, resulting in poor quantitative evaluation metrics. This issue originates from the inherent non-uniqueness associated with the principal component transformation of the method. The DBOASM method achieves incomplete cloud removal, especially in areas with slightly thicker clouds. Although the proposed PGCS$\_$M method demonstrates good spectral fidelity, its effectiveness is limited to the cirrus clouds. PGCS$\_$M is unable to correct the non-cirrus clouds because its estimation of multi-channel cloud is totally reliant on the cirrus band. In contrast, both GTCCM and the proposed PGCS$\_$D method effectively remove all the clouds, while maintaining favorable spectral characteristics. The cloud correction result of the proposed PGCS$\_$D method demonstrates the highest global consistency, as illustrated in Fig. \ref{fig:fig13}h, and PGCS$\_$D also achieves the best quantitative results.

\begin{table}
    \centering
\caption{\label{tab:table4}Evaluation of the different correction methods in a non-cirrus thin cloud covered region.}
    \begin{tabular}{>{\centering\arraybackslash}p{3.5cm}>{\centering\arraybackslash}p{1.5cm}>{\centering\arraybackslash}p{1.5cm}>{\centering\arraybackslash}p{1.5cm}>{\centering\arraybackslash}p{1.5cm}>{\centering\arraybackslash}p{1.5cm}}
    \toprule
         &  PSNR↑&  SSIM↑&  CC↑&  SAM↓& RMSE↓\\
         \midrule
         CR-NAPCT&  13.0598 &  -0.6585 &  0.6988 &  157.3943 & 0.2223 
\\
         DBOASM&  34.7523 &  0.9149 &  \underline{0.8349} &  2.4790 & 0.0183 
\\
         GTCCM&  \underline{39.7194} &  \underline{0.9302} &  0.8300 &  2.4886 & \underline{0.0103} 
\\
         PGCS$\_$M& 34.8887& 0.9025& 0.7811& \underline{2.0246}& 0.0180
\\
         PGCS$\_$D&  \textbf{40.7497} &  \textbf{0.9387} &  \textbf{0.8386} &  \textbf{1.4694} & \textbf{0.0092} 
\\
         \bottomrule
         \multicolumn{6}{l}{*Best results are in \textbf{bold}, second-best results are \underline{underlined}.}\\   
    \end{tabular}
\end{table}

Overall, both PGCS$\_$M and PGCS$\_$D have the capability of cirrus cloud correction, but only PGCS$\_$D can correct non-cirrus clouds. While PGCS$\_$M may offer a slightly reduced accuracy, it is more convenient for cirrus cloud correction than PGCS$\_$D. However, the reliability of the cirrus bands places certain limitations on PGCS$\_$M. As shown in Fig. \ref{fig:fig2}, when the cirrus band image is mixed with other information, it results in incorrect correction. In addition, PGCS$\_$M cannot effectively remove clouds when the resolution of the cirrus band differs from that of the VNIR bands, as seen in Sentinel-2 data. It is also ineffective when the sensor lacks a dedicated cirrus band, as is the case with Gaofen-2 data. In practical applications, either PGCS$\_$M or PGCS$\_$D can be selected to correct cloud-contaminated images on the basis of the actual situation.

\section{Discussion}
\label{sec:5}
\setlength{\parindent}{2em}
\noindent\hspace{2em} The experiments described above have verified the accuracy and effectiveness of PGCS in cloud synthesis, as well as its support for cloud correction. However, two key issues still require further discussion. One is the comparison of a synthetic training dataset and a multi-temporal training dataset, and the other is the generation of PGCS with various sensor data.

\subsection{Comparison of synthetic dataset and multi-temporal dataset}
\label{sec:5.1}
\setlength{\parindent}{2em}
\noindent\hspace{2em} The thin cloud correction effects based on a synthetic training dataset and a multi-temporal training dataset were compared across four different deep learning networks, i.e., ResNet \cite{RN47}, UNet \cite{RN39}, SpaGAN \cite{RN70}, and FSNet \cite{RN90}. Both the synthetic dataset and the multi-temporal dataset contained the same cloud-free images. The four different deep learning networks were then trained on the two datasets. The real cloudy images and the corresponding clear images excluded from both datasets were selected to evaluate the performance of the trained models. The quantitative evaluation results for the two datasets are presented in Table \ref{tab:table5}.

\begin{figure}
\centering
\includegraphics[width=1\linewidth]{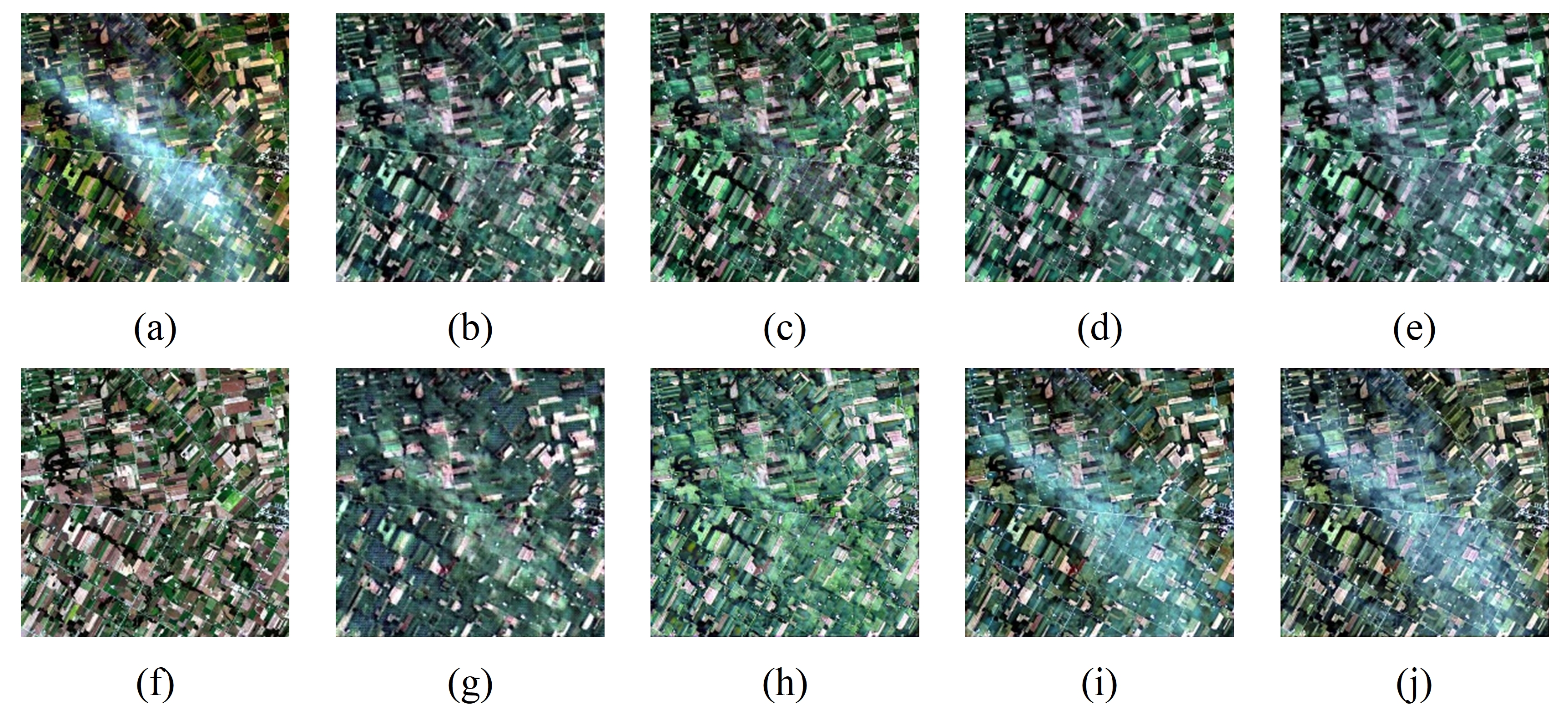}
\caption{\label{fig:fig14}Comparison of the thin cloud correction results based on a synthetic dataset (SD) and a multi-temporal dataset (MTD) under the different networks. (a) Cloudy image on September 7, 2015. (b) ResNet$\_$SD. (c) UNet$\_$SD. (d) SpaGAN$\_$SD. (e) FSNet$\_$SD. (f) Reference clear image on September 23, 2015. (g) ResNet$\_$MTD. (h) UNet$\_$MTD. (i) SpaGAN$\_$MTD. (j) FSNet$\_$MTD.}
\end{figure}

\begin{figure}
\centering
\includegraphics[width=1\linewidth]{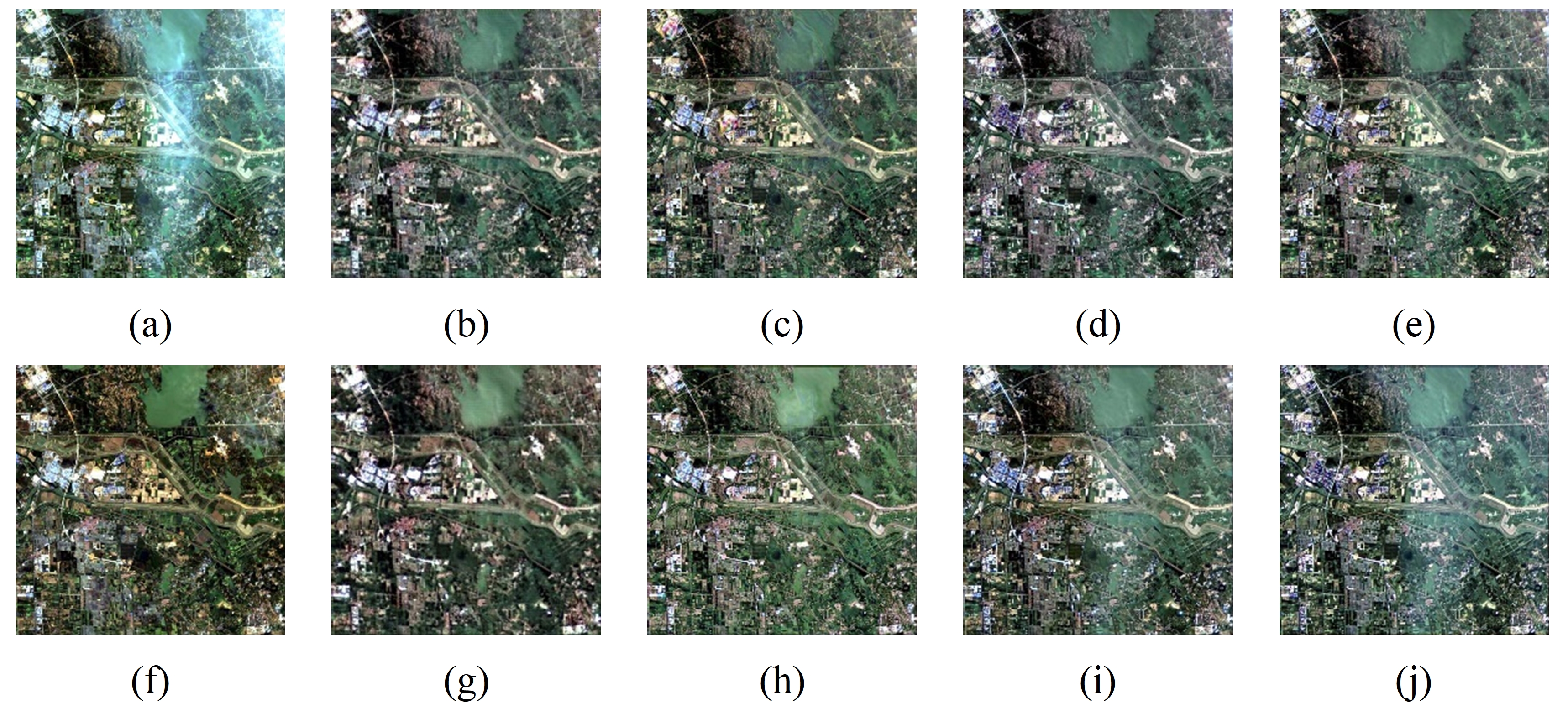}
\caption{\label{fig:fig15}Comparison of the thin cloud correction results based on a synthetic dataset (SD) and a multi-temporal dataset (MTD) under the different networks. (a) Cloudy image on September 26, 2022. (b) ResNet$\_$SD. (c) UNet$\_$SD. (d) SpaGAN$\_$SD. (e) FSNet$\_$SD. (f) Reference clear image on October 12, 2022. (g) ResNet$\_$MTD. (h) UNet$\_$MTD. (i) SpaGAN$\_$MTD. (j) FSNet$\_$MTD.}
\end{figure}

It can be observed from Table \ref{tab:table5} that FSNet trained on the synthetic dataset shows an advantage across all the quantitative metrics. By comparing the performance of the different networks on the two training datasets, it is apparent that the correction performance based on the synthetic dataset is generally superior to that of the multi-temporal dataset. The same conclusion can also be inferred from the analysis of Fig. \ref{fig:fig14}. In Fig. \ref{fig:fig14} h and i, although the clouds have been completely removed, the fidelity of the surface representation is compromised. The primary reason for this issue lies in the surface differences between the paired data acquired from adjacent time phases, which introduces uncertainties during the training of the correction model. In the design of SpaGAN and FSNet, the distinction between trained “cloudy $\&$ cloud-free” data is treated primarily as a cloud-related factor. Consequently, the surface differences lead to incomplete cloud removal in SpaGAN and FSNet, as demonstrated in Fig. \ref{fig:fig14} h and j. Meanwhile, PGCS ensures consistent surface information for the paired data, which leads to more stable and accurate cloud correction results. Additional correction examples are shown in Fig. \ref{fig:fig15}.

PGCS can effectively create a synthetic paired “cloudy $\&$ cloud-free” dataset, reducing the cost of multi-temporal paired data collection and improving the correction effect of deep learning networks. These findings further highlight the importance of PGCS in the subsequent task.

\begin{table}
    \centering
\caption{\label{tab:table5}Evaluation of the different networks based on a synthetic training dataset and a multi-temporal training dataset.}
    \begin{tabular}{>{\raggedright\arraybackslash}p{2cm}>{\centering\arraybackslash}p{2.2cm}>{\centering\arraybackslash}p{1.8cm}>{\centering\arraybackslash}p{1.8cm}>{\centering\arraybackslash}p{1.8cm}>{\centering\arraybackslash}p{1.8cm}>{\centering\arraybackslash}p{1.8cm}}
    \toprule
          Dataset&Network&  PSNR↑&  SSIM↑&  CC↑&  SAM↓& RMSE↓\\
         \midrule
         \multirow{4}{2cm}{Synthetic} &ResNet&  34.2852 &  0.8350 &  0.7099 &  3.6999 & 0.0194 \\
         &UNet&  33.5056 &  0.7976 &  0.7175 &  3.3409 & 0.0247 \\
         &SpaGAN&  \underline{35.7101} &  \underline{0.8981} &  \underline{0.8266} &  \underline{3.0566} & \underline{0.0165} \\
         & FSNet& \textbf{35.9253} & \textbf{0.8996} & \textbf{0.8277} & \textbf{3.0352} &\textbf{0.0161} \\
         \midrule
         \multirow{4}{2cm}{Multi-temporal} & ResNet& 34.0540 & 0.8425 & 0.7353 & 3.7676 &0.0203 \\
         &UNet& 35.0844 & 0.8835 & 0.8066 & 3.4254 &0.0180 \\
         &SpaGAN& 35.2757 & 0.8849 & 0.8079 & 3.3205 & 0.0175 \\
         &FSNet&  35.2356 &  0.8831 &  0.8034 &  3.2614 & 0.0175 \\
         \bottomrule
          \multicolumn{7}{l}{*Best results are in \textbf{bold}, second-best results are underlined.}\\   
    \end{tabular}
\end{table}

\subsection{Generalization to various sensors}
\label{sec:5.2}
\setlength{\parindent}{2em}
\noindent\hspace{2em} The proposed PGCS method was applied to Sentinel-2 and GaoFen-2 data to verify its general applicability. Due to the differences in band settings and data acquisition conditions between the different sensors, the parameter settings of the proposed method were adjusted accordingly. The single-channel thin cloud in the PGCS method was generated based on the cirrus band data from Landsat-8/9. Thus, a central wavelength of 1.375 µm was selected for the data generated by the cloud spatial synthesis model. In the cloud spectral synthesis phase, different central wavelengths were employed for the data simulation according to the specific bands of each sensor, as detailed in Table \ref{tab:table6}. In addition, the maximum values of the channel offset were set based on the characteristics of the different sensors, with Landsat-8/9 assigned a value of 2, Sentinel-2 set to 5, and Gaofen-2 set to 0.

\begin{table}
    \centering
\caption{\label{tab:table6}Central wavelengths of the different sensors across different bands.}
    \begin{tabular}{>{\centering\arraybackslash}p{2cm}>{\centering\arraybackslash}p{1.5cm}>{\centering\arraybackslash}p{1.5cm}>{\centering\arraybackslash}p{1.5cm}>{\centering\arraybackslash}p{1.5cm}>{\centering\arraybackslash}p{1.5cm}}
    \toprule
         &  Coastal band&  Blue band&  Green band &  Red band& NIR band\\
         \midrule
         Landsat-8/9&  0.4500&  0.4626&  0.5613&  0.6546& 0.8650\\
         Sentinel-2&  0.4430&  0.4900&  0.5600&  0.6650& 0.8420\\
         Gaofen-2&  /&  0.4850&  0.5550&  0.6600& 0.8330\\
         \bottomrule
          \multicolumn{6}{l}{*Units of wavelengths: micrometers (µm).}\\   
    \end{tabular}
\end{table}

\begin{figure}
\centering
\includegraphics[width=1\linewidth]{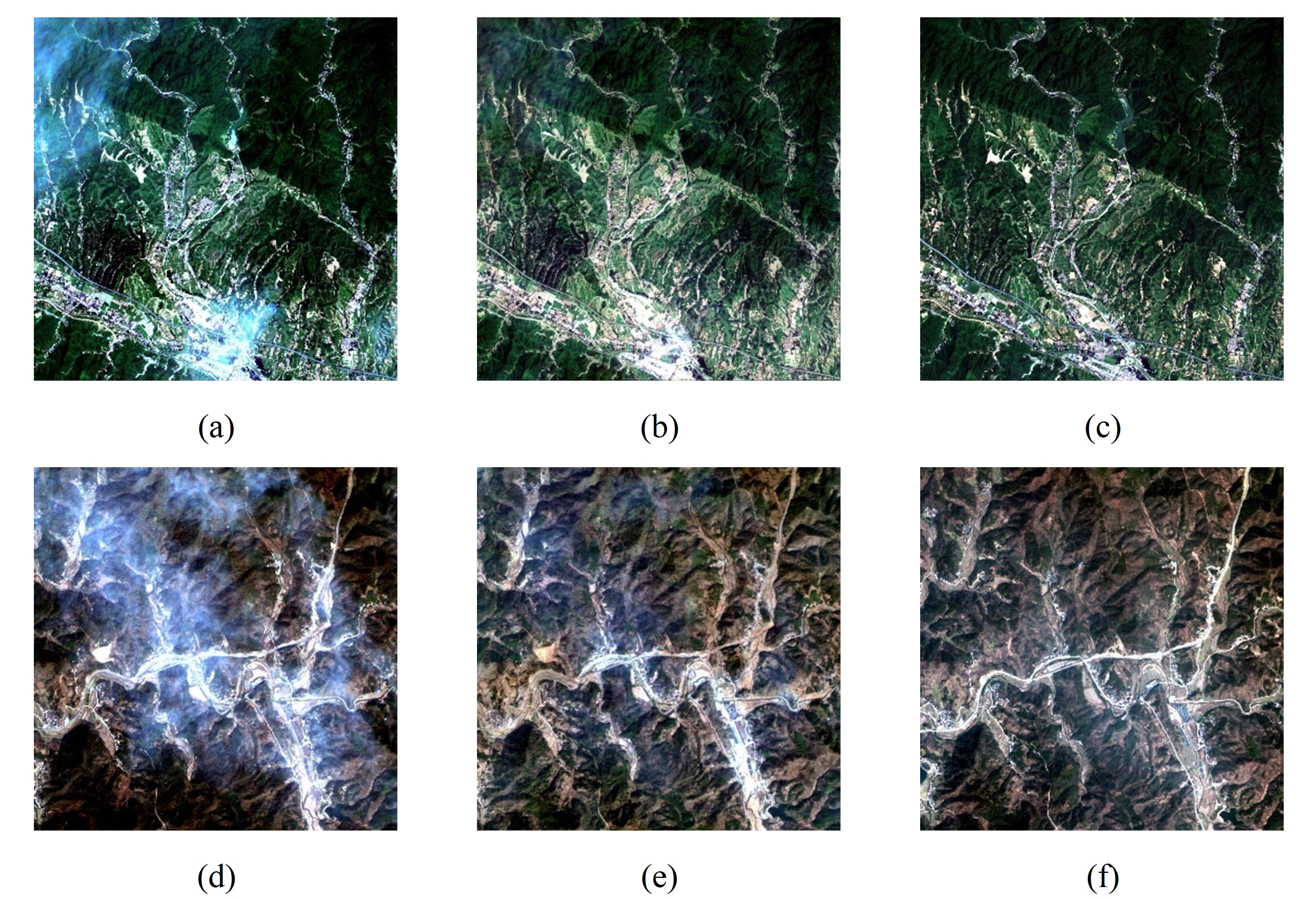}
\caption{\label{fig:fig16}Thin cloud correction results based on the proposed synthesis method with different sensors. (a) Sentinel-2 cloudy image on June 9, 2023. (b) Correction result for Sentinel-2. (c) Sentinel-2 clear image on June 24, 2023. (d) Gaofen-2 cloudy image on November 6, 2021. (e) Correction result for Gaofen-2. (f) Gaofen-2 clear image on February 3, 2021.}
\end{figure}

Fig. \ref{fig:fig16}a presents the cloudy Sentinel-2 data acquired on June 9, 2023, with a central coordinate of \ang{33;3;51.29}N, \ang{111;9;11.29}E. Fig. \ref{fig:fig16}b shows the correction result of PGCS$\_$D. Fig. \ref{fig:fig16}c depicts a clear reference image of the same area, acquired on June 24, 2023. Fig. \ref{fig:fig16}d illustrates the cloudy Gaofen-2 data acquired on November 6, 2021, with a central coordinate of \ang{31;31;19.95}N, \ang{114;55;13.38}E. Fig. \ref{fig:fig16}e displays the correction result of PGCS$\_$D, while Fig. \ref{fig:fig16}f presents the corresponding clear reference image acquired on February 3, 2021. The cloud correction results indicate that the proposed PGCS method exhibits strong generalization capabilities, making it highly effective for cloud simulation and cloud degradation dataset construction across various satellite data, including Landsat-8/9, Sentinel-2, and Gaofen-2. This ability could provide valuable support for subsequent tasks with various sensor data, highlighting its practical application value.

\section{Conclusions}
\label{sec:6}
\setlength{\parindent}{2em}
\noindent\hspace{2em} In this paper, we have presented a novel cloud synthesis method named PGCS that effectively integrates data-driven algorithms, physical models, and statistical information. The PGCS method enhances cloud shape diversity by leveraging a generative adversarial network, while accurately representing the spectral radiation difference of cloud in the different bands through physical models and statistical information.

The proposed PGCS method can synthesize infinite cloud images with high fidelity, which can be used to construct paired datasets consisting of cloud, ground and cloudy images. These datasets can effectively support the cloud correction task. The experiments conducted in this study also demonstrated that this data-driven method significantly outperforms the traditional statistical methods in cloud correction. In addition, the performance of the synthetic training dataset was found to more stable than that of a multi-temporal training dataset. Furthermore, this method can be extended to various sensors and is suitable for remote sensing images cloud processing tasks excluding the cirrus band. It is also expected to support other cloud-related tasks and serve as a data augmentation method to enhance the performance of algorithms on low-quality images.

Today is the era of big data and artificial intelligence. Data-driven algorithms provide us with greater possibilities. With the utilization of generative adversarial networks, we can not only learn distribution features from data but also enhance the diversity and variability of the data. Secondly, physical models still hold irreplaceable advantages in describing Earth surface processes. We firmly believe that the coupling of these two representative frameworks will greatly contribute to the further advancement of geoscience.

\section*{Acknowledgments}
\setlength{\parindent}{2em}
\noindent\hspace{2em} This work was funded by the National Key Research and Development Program of China (2022YFF1301103), the National Natural Science Foundation (42422109), and the Fundamental Research Funds for the Central Universities (2042023kfyq04). The authors would like to thank all the researchers who kindly shared the codes used in this paper.

\bibliographystyle{unsrt}
\bibliography{sample}

\end{document}